%% file: paper.tex
\documentclass[sigconf]{acmart}

\usepackage{hyperref}       % hyperlinks
 \hypersetup{
     colorlinks=true,
     linkcolor=blue,
     filecolor=blue,
     citecolor=red,      
     urlcolor=cyan,
     }

\makeatletter
\newcommand*{\rom}[1]{\expandafter\@slowromancap\romannumeral #1@}
\makeatother
\usepackage{xspace}
\usepackage{bbm}
\usepackage{tabularx, booktabs}
\newcolumntype{C}[1]{>{\centering\let\newline\\\arraybackslash\hspace{0pt}}m{#1}}
\newcolumntype{R}[1]{>{\raggedleft\let\newline\\\arraybackslash\hspace{0pt}}m{#1}}
\newcolumntype{L}[1]{>{\raggedright\let\newline\\\arraybackslash\hspace{0pt}}m{#1}}
\usepackage{multirow}
\usepackage{array}
\usepackage{subfig}
\usepackage{graphicx}
\usepackage{amsmath}
\usepackage{mathtools}
\usepackage{amsthm}
\usepackage{algorithm}
\usepackage{algorithmic}
\usepackage{titlesec}

\titlespacing*{\subsection}{5pt}{0.15\baselineskip}{0.2\baselineskip}
\titlespacing*{\subsubsection}{5pt}{0.1\baselineskip}{0.1\baselineskip}
\titlespacing*{\section}{5pt}{0.5\baselineskip}{0.5\baselineskip}

\input{dfn}

\acmConference[WWW '24]{}{May 13--17, 2024}{Singapore}
\title{Descriptive Kernel Convolution Network with Improved Random Walk Kernel}

\author{Meng-Chieh Lee}
\authornote{Both authors contributed equally to this research.}
\affiliation{
  \institution{Carnegie Mellon University}
  \city{Pittsburgh}
  \state{PA}
  \country{USA}
}
\email{mengchil@cs.cmu.edu}

\author{Lingxiao Zhao}
\authornotemark[1]
\affiliation{
  \institution{Carnegie Mellon University}
  \city{Pittsburgh}
  \state{PA}
  \country{USA}
}
\email{lingxiao@cmu.edu}

\author{Leman Akoglu}
\affiliation{
  \institution{Carnegie Mellon University}
  \city{Pittsburgh}
  \state{PA}
  \country{USA}
}
\email{lakoglu@andrew.cmu.edu}

\begin{document}

\begin{abstract}
\input{000abstract}
\end{abstract}



%%
%% Keywords. The author(s) should pick words that accurately describe
%% the work being presented. Separate the keywords with commas.


\maketitle

\setlength{\abovedisplayskip}{5pt}
\setlength{\belowdisplayskip}{5pt}

\section{Introduction}
\label{sec:intro}
\input{010introduction}
\section{Related Work} \label{sec:background}
% \vspace{-0.5mm}
\input{020background}

% \vspace{-3mm}
\section{Kernel Convolution Networks with Random Walk Kernel and Beyond} \label{sec:rw_kernel}
% \vspace{-1mm}
\input{030RWKernel.tex}

% \vspace{-2mm}
\section{Experiments \rom{1}: Unsupervised Pattern Mining} \label{sec:part1}
% \vspace{-1mm}
\input{040exp-part1.tex}

% \section{Experiments \rom{2}: Adapting to Various Architectures and Applications}
\section{Experiments \rom{2}: Adapting to Various Applications}
\label{sec:part2}
\input{050exp-part2.tex}

\section{Conclusion}
\label{sec:concl}
\input{060conclusion}

\clearpage
\balance
\bibliographystyle{unsrt}
\bibliography{ref.bib}

\clearpage
\appendix

\input{070appendix}

\end{document}

%% file: dfn.tex
\newcommand\sbullet[1][.5]{\mathbin{\vcenter{\hbox{\scalebox{#1}{$\bullet$}}}}}

\newtheorem{definition}{Definition}

\newcommand{\cbit}{\begin{compactitem}}
\newcommand{\ceit}{\end{compactitem}}
\newcommand{\cben}{\begin{compactenum}}
\newcommand{\ceen}{\end{compactenum}}

\newcommand{\bit}{\begin{compactitem}}
\newcommand{\eit}{\end{compactitem}}
\newcommand{\ben}{\begin{compactenum}}
\newcommand{\een}{\end{compactenum}}

\newcommand{\kername}{\textsc{RWK$^+$}\xspace}
\newcommand{\method}{\textsc{RWK$^+$CN}\xspace}
\newcommand{\colormatch}{color-matching\xspace}
\newcommand{\structure}{structural colors\xspace}
\newcommand{\diversity}{diversity\xspace}
\newcommand{\rwk}{RWNN\xspace}

\newcommand{\hiddens}{hidden graphs\xspace}
\newcommand{\hidden}{hidden graph\xspace}
\newcommand{\norm}{\textsc{StepNorm}\xspace}

\newcommand{\rwori}{RWK\xspace}
\newcommand{\rwpp}{RWK$^+$\xspace}
\newcommand{\rwppc}{RWK$^+$Conv\xspace}

\newcommand{\ckn}{\textsc{KCN}\xspace}
\newcommand{\ckns}{\textsc{KCN}s\xspace}

\newcommand{\codeurl}{\url{https://github.com/mengchillee/RWK_plus}}

\newcommand{\emphasize}[1]{\textbf{\underline{\smash{#1}}}}

%% file: 000abstract.tex
% Weisfeiler-Leman (WL) and random walk kernels have been the two most widely used graph kernels for graph feature engineering before the surge of GNNs. However, graph kernels are quickly superseded by GNNs as the former lacks learnability. GNNs are closely related to the WL test \cite{weisfeiler1968leman} and the WL kernel \cite{hido2009linear}. 

{Graph kernels used to be the dominant approach to feature engineering for structured data, which are superseded by modern GNNs as the former lacks learnability. Recently, a suite of Kernel Convolution Networks (\ckns) %\cite{chen2020convolutional} 
successfully revitalized graph kernels by introducing learnability, which convolves input with learnable \hiddens using a certain graph kernel.}
The random walk kernel (RWK)
% , which is historically the first graph kernel \cite{gartner2003graph,kashima2003marginalized}, 
has been used as the default kernel in many \ckns, gaining increasing attention. 
% In this paper, we first show that \ckns can be adapted to the unsupervised setting toward learning descriptive graph features that cannot be achieved by GNNs.
In this paper, we first revisit the RWK and its current usage in \ckns, revealing several shortcomings of the existing designs, and propose an improved graph kernel \kername, by introducing \colormatch random walks and deriving its efficient computation.
We then propose \method, a KCN that uses \kername as the core kernel to learn descriptive graph features with an unsupervised objective, which can not be achieved by GNNs.
% We improve the design by introducing \colormatch random walks, and show that the unrolling of the proposed version for efficient computation shares connection to GNNs. 
% We capitalize on this connection to employ \norm, a combination of BatchNorm and sigmoid for learnable similarity normalization. 
% Further, we enrich node labels with ``\structure'' (i.e. node embeddings) and introduce diversity regularization toward capturing frequent and distinct graph patterns. 
Further, by unrolling \kername, we discover its connection with a regular GCN layer, and propose a novel GNN layer \rwppc.
In the first part of experiments, we demonstrate the descriptive learning ability of \method with the improved random walk kernel \kername on unsupervised pattern mining tasks;
in the second part, we show the effectiveness of \kername for a variety of \ckn architectures and supervised graph learning tasks, and demonstrate the expressiveness of \rwppc layer, especially on the graph-level tasks.
\kername and \rwppc adapt to various real-world applications, including web applications such as bot detection in a web-scale Twitter social network, and community classification in Reddit social interaction networks.\looseness=-1

%% file: 010introduction.tex
%%%%%%%%%%%%%%%%%%%%%%%%%%%%%%%%%%%%%%%
% AUTHOR: Christos Faloutsos
% INSTITUTION: CMU
% DATE: April 2019
% GOAL: to streamline the paper presentations
%%%%%%%%%%%%%%%%%%%%%%%%%%%%%%%%%%%%%%%

%\notice{Again, a rhetorical question}

% 1st order: learnable structures / kernel similarity
% related: a) flipping the objective from supervised to unsupervised; b) unsupervised graph embedding
% a) makes it descriptive instead of discriminative
% discriminative can be useful / solve the task

{Graph kernels have historically been a popular approach to ``flatten'' graphs explicitly or implicitly into vector form that many downstream algorithms can more easily handle. 
% The large body of graph kernels include those based on shortest paths \cite{borgwardt2005shortest}, subtrees \cite{mahe2009graph}, graphlets \cite{shervashidze2009efficient}, random walks \cite{gartner2003graph}, and others (see \cite{nikolentzos2021graph,kriege2020survey} for survey). 
While graph kernels exhibit mathematical expressions that lend themselves to theoretical analysis \cite{gartner2003graph}, their handcrafted features may not be expressive enough to capture the complexities of various learning tasks on graphs \cite{ramon2003expressivity}. 
More recently, graph kernels are superseded by modern GNNs which leverage multi-layer architecture and nonlinear transformations to learn task-adaptive graph representations \cite{zhou2020graph}.}

Interestingly, GNNs bear a close connection to the Weisfeiler-Leman (WL) graph kernels \cite{shervashidze2011weisfeiler}, as well as the related WL graph isomorphism test \cite{weisfeiler1968leman}. 
In fact, most recent work on the expressive power of GNNs heavily use the $k$-WL hierarchy \cite{leskovec2019powerful,sato2020survey}, and others have derived inspiration from it to design novel GNN architectures \cite{morris2019weisfeiler,maron2019provably,bouritsas2022improving,zhao2022practical}.
%In essence 
The WL kernel, which is quite popular thanks to its attractive linear-time complexity \cite{hido2009linear}, derives its simplicity from iterative neighborhood aggregation, akin to the convolution scheme of message-passing GNNs \cite{gilmer2017neural}. 
This type of connection has been recognized and leveraged in the recent few years to derive a series of ``\textit{GNNs meet graph kernels}'' style models that bridge these two worlds \cite{mairal2016end,lei2017deriving,chen2020convolutional,du2019graph,nikolentzos2020random,cosmo2021graph,feng2022kergnns,al2019ddgk,long2021theoretically} (see Sec.~\ref{sec:background} for detailed related work), named as Kernel Convolutional Networks (KCNs). 
% \looseness=-1

% *** RQ: *** Given a graph database, how can we learn graph representations that are reflective of the random walk graph kernel similarities, which are efficiently learnable end-to-end, rather than hand-crafted by a pre-specified expression? 

% \leman{why random walk kernel?}
% ---> The first methods for graph comparison referred to as graph kernels were proposed in 2003 [22, 23]
% ----> Gärtner et al. [22] and Kashima et al. [23] simultaneously proposed graph kernels based on random walks, which count the number of (label sequences along) walks that two graphs have in common. 
% ---> Random walk kernels [12, 21] have been the starting point of a long line of research in graph kernels

% In this paper, we deepen the synergy between GNNs and graph kernels by proposing a \textit{descriptive} kernel convolution network that utilizes an \textit{improved random walk kernel} called \kername. 
RWKs, based on the number of (node label sequences along) walks that two graphs have in common, have been the starting point in the history of graph kernels \cite{gartner2003graph,kashima2003marginalized}.
% \cite{kriege2020survey}, dating back to 2003 
Notably, a recent study by \cite{kriege2022weisfeiler}
demonstrated that classical random walk kernels with only minor modifications are as expressive as the Weisfeiler-Leman kernels and even surpass their accuracy on real-world classification tasks.
% "Given a graph kernel of choice, in each GKC layer the input graph is compared against a series of structural masks (analogous to the convolutional masks in CNNs), which effectively represent learnable sub-graphs. These in turn can offer better intrepretability, in the form of insights into what structural patterns in the input graphs are related to the corresponding output."
Inspired by this connection, our work extends from the random walk neural network (RWNN) of \cite{nikolentzos2020random}, 
where each input graph is represented by its RWK similarity to a set of small graphlets (called \hiddens) that are learned end-to-end by optimizing a classification objective. 

In this paper, we deepen the synergy between GNNs and graph kernels, and improve the RWK as utilized within GNNs in a number of fronts. 
First, toward capturing more representative patterns, we introduce several improvements to the RWK in both effectiveness and efficiency and propose an improved graph kernel \kername. 
Second, we propose a \textit{descriptive} KCN \method by flipping the objective from a {discriminative} one to a {descriptive} one that helps us capture relational patterns in the graph database. 
What is more, we derive the mathematical connection of \kername to layer-wise neural network operators for the first time, which inspires us to propose a novel GNN layer \rwppc. 
Finally, we employ our \kername and \rwppc on a suite of real-world tasks for graph data and achieve significant gains.
% , including anomaly detection, substructure counting and graph classification, achieving significant gains. 
A summary of our contributions is as follows:\looseness=-1

% within our final proposed kernel \kername. 
% These so-called structural colors provide further characterizing information for nodes along a walk and help better capture structural similarity between graphs.

% $\sbullet[.75]$ \textbf{RWK with efficient color-matching and learnable normalization:} 
$\sbullet[.75]$ \textbf{\kername with efficient color-matching:} 
We identify that the RWK originally developed in RWNN only enforces the same label at the start and the end of two walks while ignoring the intermediates. We reformulate it to count a walk as shared only if all corresponding node pairs exhibit the same node label (i.e. color) at all steps along the walk. %, called RWK-\textbf{CM} for RWK with ``\textbf{c}olor-\textbf{m}atching''. 
While more reflective of graph similarity, RWK with color-matching incurs a memory and computational overhead. 
Therefore, we propose the improved graph kernel \kername through transforming its formulation for efficient computation. 
In addition, we propose a learnable solution \norm to address the nontrivial task of combining similarity scores across steps with drastically different scales.\looseness=-1
% To this end, we propose \norm that normalizes each step in a learnable fashion, avoiding handcrafted weights for each step's similarity. 

% Finally, we address the nontrivial task of combining similarity scores across different RWK steps with drastically different scales. To this end, we propose \norm that normalizes each step in a learnable fashion, avoiding handcrafted weights for each step's similarity. 

% $\sbullet[.75]$ \textbf{Enhancing RWK further with ``structural colors'':} Besides the node labels in labeled graphs as well as for unlabeled graphs, we use additional structural features associated with each node within our final proposed kernel \kername. These so-called structural colors provide further characterizing information for nodes along a walk and help better capture structural similarity between graphs.

$\sbullet[.75]$ \textbf{\method learning descriptive hidden graphs}: 
The original RWNN is trained supervised for graph classification and thus learns {discriminative} \hiddens. 
We propose \method with an unsupervised objective, that uses \rwpp as the core kernel and maximizes the total RWK similarity between the input graphs and \hiddens.
The learned \hiddens are reflective of the frequent walks (i.e. patterns) in the database.
To further enhance the descriptive ability, we use additional ``structural colors'' to help better capture structural similarity between graphs, and enforce a {diversity regularization} among the \hiddens to capture non-overlapping subgraphs.
% associated with each node to provide further characterizing information for nodes along a walk and
% with distinct frequent walk patterns.
Finally, we demonstrate the descriptive learning ability of \method with our carefully designed testbeds.

$\sbullet[.75]$ \textbf{\rwppc, a novel GNN layer}: 
By unrolling \rwpp, we discover that the derivation can be re-written as a sequence (i.e. multiple layers) of graph convolutional operations, connecting with regular GCN layers. 
By viewing hidden graphs as learnable parameters, we transform the \rwpp algorithm into a novel GNN layer called \rwppc. 
The \rwppc layer uses additional element-wise product operation that can potentially bring better expressiveness than the GCN layer. 
% Empirically, it outperforms GCN in both node- and graph-level tasks, but by a large margin in graph-level tasks.  

$\sbullet[.75]$ \textbf{Broad applications of \kername and \rwppc}: 
% Our \kername is an improved version of vanilla RWK in both efficiency and efficacy, yet, is independent of the objective being used or any particular KCN.
% To showcase the advantages of \kername over the vanilla RWK, 
We employ \kername as the core kernel inside different KCN architectures and evaluate it on four graph-level tasks: one discriminative (graph classification), and three descriptive (graph pattern mining, graph-level anomaly detection, and substructure counting).
It is shown to be improved over the vanilla RWK especially on descriptive tasks.
Moreover, we compare our proposed \rwppc layer with the GCN layer on node- and graph-level tasks.
\rwppc outperforms GCN in both tasks, notably by a large margin in graph-level tasks, empirically demonstrating its better expressiveness.  
It is worth noting that our experiments contain a broad-range of real-world applications, including web applications such as bot detection in a web-scale Twitter social network with a million nodes, and community classification in Reddit social interaction networks.

% We improve over the vanilla RWK especially on descriptive task performance.
% where anomaly detection can be seen as the flip-side of pattern mining, and substructure counting is informed by frequent pattern matching.

\textbf{Reproducibility:} Our code is available at \codeurl.

%% file: 020background.tex
\textbf{Graph Kernels.}
The literature on graph kernels is extensive and well established, thanks to the prevalence of learning problems on graph-structured data and the empirical success of kernel-based methods \cite{nikolentzos2021graph,kriege2020survey}.
A large variety of graph kernels have been developed motivated either by their theoretical properties, or specialization or relevance to certain application domains like biology \cite{prvzulj2007biological,jie2016sub} or chemistry \cite{rupp10}.
Those include graph kernels based on shortest paths \cite{borgwardt2005shortest}, subtrees \cite{ramon2003expressivity,mahe2009graph}, graphlets \cite{prvzulj2007biological,shervashidze2009efficient}, random walks \cite{gartner2003graph,kashima2003marginalized}, as well as variants such as random walk return probabilities \cite{zhang2018retgk}, to name a few.
A long line of work focused on designing computationally tractable kernels for large graphs with discrete as well as continuous node attributes \cite{shervashidze2009efficient,conf/icml/CostaG10,conf/nips/FeragenKPBB13,DBLP:conf/icdm/MorrisKKM16}, while those such as the Weisfeiler-Leman (WL) kernel \cite{shervashidze2011weisfeiler} and others \cite{hido2009linear} gained popularity thanks to linear-time efficiency. 

A key challenge with classical graph kernels is lack of learnability; today's graph neural networks (GNNs) are able to learn feature representations that clearly supersede the fixed feature representations used by graph kernels. At the same time, several connections can be drawn between graph kernels and GNNs, such as the similarity between the neighborhood aggregation %scheme 
of the WL kernel (a.k.a. color refinement) and the scheme of message-passing GNNs \cite{gilmer2017neural}. We discuss below recent line of work that tap into the synergy between graph kernels and GNNs to harvest the best of both worlds.\looseness=-1

\textbf{Synergizing Graph Kernels and GNNs.}
While many works %that we will refer to as Convolutional Kernel Networks (\ckns) 
bridge graph kernels with GNNs, they have clear distinctions.
% in how they synergize these two worlds.
Coined as Convolutional Kernel Networks \cite{mairal2014convolutional}, and others in similar lines \cite{mairal2016end,chen2020convolutional}, introduce neural network architectures that learn graph representations that lie in the reproducing kernel Hilbert space (RKHS) of graph kernels.
Others design new classes of graph kernels using GNNs \cite{du2019graph,hazan2015steps}. 
In contrast, and closest to our work, coined very similarly as Graph Kernel Convolution Networks (\ckns) \cite{cosmo2021graph} and various others \cite{lei2017deriving,nikolentzos2020random,feng2022kergnns} integrate a graph kernel \textit{into} GNN architectures. 
In other words, they show how to realize a given graph kernel with a GNN module, which in effect unlocks end-to-end learnability for the graph kernel. 
We provide further background on \ckns in Sec.~\ref{ssec:ckn}. 
Finally, while different in focus, there is also noteworthy work exploiting graph kernels for pre-training GNNs \cite{navarin2018pre}, or to extract preliminary features that are passed onto CNNs \cite{nikolentzos2018kernel}.\looseness=-1

%bridge between kernel methods and deep networks, and ideally reach the best of both worlds.

%% file: 030RWKernel.tex
 Kernel Convolution Network (KCN) \cite{cosmo2021graph, nikolentzos2020random, feng2022kergnns} that convolves the input graph with learnable \hiddens using a certain graph kernel has gained 
increasing attention recently, as it offers learnability to graph kernels. Given the simplicity of random walk kernel (RWK) and its differentiability, it has been used as the default graph kernel in many \ckns like RWNN \cite{nikolentzos2020random} and KerGNN \cite{feng2022kergnns}. 
We first introduce notation and background of KCN, along with designing an unsupervised loss for learning descriptive features (Sec. \ref{ssec:ckn}). %(Sec. \ref{sssec:descriptive}). 
Then we revisit the RWK (Sec. \ref{ssec:rw}), and discuss the issues of its current usage in \ckns (Sec. \ref{ssec:issues}). Next, we introduce color-matching based RWK, along with its efficient computation that shares connection to GNNs (Sec. \ref{ssec:color-matching}). Finally, we discuss how to increase the descriptive ability of the learned hidden graphs in the unsupervised setting (Sec. \ref{ssec:diversity}). \looseness=-1

\textbf{Notation:} Let $G$$=$$(V(G), E(G), l_G)$ denote an undirected, node-attributed graph with $n$ nodes in $V(G)$, $e$ edges in $E(G)$, and an attribute or labeling function $l_G: V(G)\rightarrow C$ where $C$ can be $\mathbb{R}^d$ for %$d$-dimensional 
continuous attributes or $\{c_1,...,c_d\}$ for %$d$ 
distinct discrete labels. Let $A_G$ denote the adjacency matrix, %representation of $E(G)$, 
and $ A_{G\otimes H}:= A_G \otimes A_H $ depict the Kronecker product of the adjacency matrices for graphs $G$ and $H$. 
Let $X_G:=[\mathbf{x}_{v_1}, \ldots, \mathbf{x}_{v_{n}} ]^T \in \mathbb{R}^{n \times d}$ be the node attributes in $G$. 

% \vspace{-3mm}
\subsection{Kernel Convolution Networks}
\label{ssec:ckn}
% \vspace{-1mm}
% \textcolor{red}{I thought CKNs are more than these 3, but I was wrong, this can hurt the contribution. In the meanwhile, the term convolutional kernel network has been used for describing a different things, we should change the term. }

Graph kernels are designed to measure similarity on a pair of graphs. However, they produce fixed handcrafted features. \cite{lei2017deriving} derived the first neural network that outputs the RWK similarity scores between input graph and hidden learnable path-like graphs. \cite{nikolentzos2020random} generalized \cite{lei2017deriving} such that the hidden graphs can have any structure without the path constraints. 
The designed model is claimed to be interpretable as the learned hidden graphs ``summarize'' the input graphs. Later, \cite{cosmo2021graph} and \cite{feng2022kergnns} extended RWNN \citep{nikolentzos2020random} to a multi-layer architecture, in which each layer compares subgraphs around each node of the input with learnable hidden graphs. We refer to these models as Kernel Convolution Networks (\ckns) as they generalize the Convolutional Neural Network (CNN) from the image domain to the graph domain, with the help of a graph kernel. 
Each layer of the KCN has a number of learnable \hiddens. 

Formally, let $G$ be the input graph with node $v \in V(G)$; let $\mathbf{h}^t(v) \in \mathbb{R}^{k_t} $ be the representation of node $v$ at the $t$-th layer where $k_t$ is the number of learnable kernels in KCN's $t$-th layer for $t>0$, and $k_0$ be the dimension of original node attributes with $\mathbf{h}^0(v) = \mathbf{x}_v$. Let $W^t_1,..., W^t_{k_t}$ denote the series of learnable hidden graphs in the $t$-th layer, and $\text{Sub}^{t}_G[v]$ be the subgraph around node $v$ on $G$ with attributes $\{\mathbf{h} ^t(u) | u \in \text{Sub}_G[v]\}$, we have:
% \textbf{Multi-layer KCN}
\begin{align}
    \mathbf{h}^{t+1}(v) = [ \mathcal{K}(\text{Sub}^{t}_G[v], W^{t+1}_1), \ldots,  \mathcal{K}(\text{Sub}^{t}_G[v], W^{t+1}_{k_{t+1}}) ] \;,
\end{align}

\noindent where $\mathcal{K}$ is the graph kernel used to compute graph similarity. Multi-layer \ckns stack graph kernel computations with layers, and output node representations at each layer which can be used for any downstream task. They exhibit strong representation ability however the output is not interpretable or descriptive. The single-layer \ckn, while less expressive, can output meaningful similarity scores for descriptive unsupervised feature learning, which % Formally, a single-layer \ckn 
computes graph-level representation directly by:
\begin{align}
    \mathbf{h}(G) = [ \mathcal{K}(G, W_1), \dots, \mathcal{K}(G, W_{k}) ] \;.
\end{align}

\noindent\textbf{Learning Descriptive Features}.
\ckns were originally proposed for supervised learning, as such, the learned \hiddens are discriminative for classification tasks. We claim that the KCN model can be paired with an unsupervised loss and used to generate \textit{descriptive} \hiddens instead, which is not achieved by existing GNNs. Given the output of a single-layer \ckn model is the similarity scores to each hidden graph, one can train the \ckn by maximizing the total similarity score.
% (as opposed to minimizing the classification loss). 
Specifically, the unsupervised objective is given as:\looseness=-1
{
\begin{align}
  \max_{W_{1}, \dots, W_{k}}{\sum_{i=1}^{k}{\mathcal{K}(G, W_{i})}} \;. \label{eq:unsupervised}
\end{align}
}
% $\max_{~W_{1}, \dots, W_{m}}{\sum_{i=1}^{m}{\mathcal{K}(G, W_{i})}}$.

\noindent
With this new objective, the learnable hidden graphs are to reflect or summarize the common patterns of the graph database.
Put differently, similarities are maximized when the learned graphs capture frequent structural patterns that the kernel is designed to capture.\looseness=-1
%Hence we can use the output similarity score and corresponding hidden graph to describe the feature. %\textcolor{red}{maybe a bit weak, not sure how to write clearly}

\subsection{Revisiting the Random Walk Kernel (RWK)}
\label{ssec:rw}
RWK has been used in KCNs as the default kernel. It has been originally proposed to compare two labeled graphs by counting the number of common walks on both graphs \cite{gartner2003graph,kashima2003marginalized}. Formally, consider a labeled (discrete attribute) graph $G$ %with labeling function $l$ 
such that $l(v)$ represents the label of node $v \in V(G)$. Let $\mathcal{R}_{t}(G)$ be the set of all $t$-step random walks on $G$. For a random walk $\mathbf{p}=(v_1, v_2, .., v_t) \in \mathcal{R}_{t}(G)$, let $l(\mathbf{p}) = (l(v_1), ..., l(v_t))$ denote the labels along the walk. 
Then the $t$-step RWK $\mathcal{K}^{t}_{rw} (G, H)$ computes the similarity of $G$ and $H$ by counting the common walks as follows:
{\small
\begin{align}
 \mathcal{K}^{t}_{rw} (G, H) = \sum_{i=1}^t \lambda_i \sum_{\mathbf{p} \in \mathcal{R}_i(G)}  \sum_{\mathbf{q} \in \mathcal{R}_i(H)} \text{I}\big(l(\mathbf{p}), l(\mathbf{q})\big) \label{eq:rw1}
\end{align}
}

\noindent
where $\text{I}(x,y)$ is the Dirac kernel where $\text{I}(x,y)=1$ if $x=y$, and 0 otherwise; and $\lambda_i \in \mathbb{R}$ denotes the weight of the $i$-th step's score. 
\begin{definition}(Direct graph product)
Given two labeled graphs $G,H$ with labeling function $l$, their direct product is a new graph $G \times H$ with adjacency matrix $A_{G\times H}$, vertices $V(G \times H) = \{ (u, v) \in V(G) \times V(H)\  | \ l(u)= l(v) \}$ and edges $E(G \times H) = \{\big((u_1, v_1), (u_2, v_2)\big) \in V^2(G\times H) \ |\ (u_1, u_2) \in E(G)\ {\text{and}} \ (v_1, v_2) \in E(H)  \}. $
\end{definition}

\cite{gartner2003graph} have shown that for any length $t$ walk, there is a bijective mapping between $\mathcal{R}_t(G \times H)$ and $\{ (\mathbf{p}, \mathbf{q}) \in  \mathcal{R}_t(G) \times  \mathcal{R}_t(H) \ |\ l(\mathbf{p}) = l(\mathbf{q}) \}$. Therefore, Eqn.~\eqref{eq:rw1} can be rewritten as:
{\small
\begin{align}
    \mathcal{K}^{t}_{rw} (G, H) = \sum_{i=1}^t \lambda_i %\revise{\cdot}  
    (\mathbf{1}^T A_{G\times H}^i \mathbf{1}) \label{eq:rw2}
\end{align}
}
\noindent where $\mathbf{1}$ denotes the all-ones vector of length $|V(G\times H)|$. Note that $A_{G \times H}$ is \textit{not} $A_{G\otimes H}$, where the latter is the Kronecker product of %the adjacency matrices 
$A_G$ and $A_H$ without enforcing label-matching along the walk. 

\subsection{Issues of Adapting RWK to \ckn} 
\label{ssec:issues}
% Maximizing \rwk without node features equals maximizing the number of edges in the learned graphlets.
% Maximizing \rwk with node features proposed by RWGNN equals maximizing the $d \times d$ matrix.

The original RWK is designed for labeled graphs and cannot handle graphs with continuous node attributes KCNs are often used for. To that end, \cite{nikolentzos2020random} proposed an extension of the RWK in their RWNN. % to handle continuous-attributed graphs. 
Let $X_G \in \mathbb{R}^{|V(G)| \times d}$ depict the $d$-dimensional continuous attributes for all nodes, and $X_H \in \mathbb{R}^{|V(H)| \times d}$ and $A_H \in \mathbb{R}^{|V(H)| \times |V(H)|}$ depict the learnable node features and the learnable adjacency matrix of the hidden graph $H$, respectively. For two graphs $G$ and $H$, let $S = X_H X_G^T \in \mathbb{R}^{|V(H)| \times |V(G)|}$ encode the dot product similarity between the attributes of the vertices from two graphs, where $\mathbf{s}:= \text{vec}(S)$ is the 1-d vectorized representation of $S$. The authors of RWNN proposed to compute the revised RWK as:
\begin{align}
    \mathcal{K}^{t}_{rw-} (G, H) = \mathbf{1}^T ( \mathbf{s}\mathbf{s}^T \odot  A_{G\otimes H}^t) \mathbf{1} \label{eq:rwgnn} \;,
\end{align}
\noindent
where $\odot$ denotes the element-wise product. The revised kernel computes walks with length exactly $t$ only.
Mathematically, the term $\mathbf{s}\mathbf{s}^T$ applies reweighting to $A^t_{G\otimes H}$ such that the $(i,j)$-th element becomes $\mathbf{s}_i \mathbf{s}_j (A^t_{G\otimes H})_{ij}$ where $(A^t_{G\otimes H})_{ij}$ is equal to the number of length-$t$ walks from pair of nodes $i$ to pair of nodes $j$ in the {Kronecker product} graph $G \otimes H$. 

Although the proposed adaptation of RWK can handle continuous node attributes, we identify two critical issues with Eqn.~\eqref{eq:rwgnn} that we outline below and later address in Sec.s \ref{ssec:color-matching} and \ref{ssec:diversity}, respectively. % for both supervised end-to-end training and unsupervised descriptive feature learning. 

\textbf{Issue 1: Color mismatch.} Let $\mathbf{p} = (u_1, ..., u_t) $ be a walk on $G$ and $\mathbf{q} = (v_1,...,v_t)$ be a walk on $H$. Eqn.~\eqref{eq:rwgnn} only considers reweighting the number of walks from $(u_1,v_1)$ to $(u_t,v_t)$, where $(u_1, v_1)$ is the starting pair and $(u_t,v_t)$ the ending pair, without comparing the intermediary nodes along the walk. 
In essence, their formulation of the RWK is limited to only partially shared walks. 

\textbf{Issue 2: Inefficient parameterization.} Notice that $\mathbf{s} = \text{vec}(S) = \text{vec}(X_HX_G^T) = \sum_{i=1}^d (X_G^{[i]}\otimes X_H^{[i]})$, where $X_G^{[i]}$ denotes the $i$-th column of $X_G$. 
Using this equality, we can rewrite Eqn.~\eqref{eq:rwgnn} as:
%using the above formulation of $\mathbf{s}$ as:
{\small
\begin{align}
   \mathcal{K}^{t}_{rw-} (G, H) & = \mathbf{1}^T ( \mathbf{s}\mathbf{s}^T \odot  A_{G\otimes H}^t) \mathbf{1} = \mathbf{s}^T A^t_{G \otimes H} \mathbf{s} \nonumber \\
   & = (\sum_{i=1}^d (X_G^{[i]}\otimes X_H^{[i]}))^T (A^t_G \otimes A^t_H) (\sum_{i=1}^d (X_G^{[i]}\otimes X_H^{[i]})) \nonumber \\
   & = \sum_{i=1}^d \sum_{j=1}^d (X_G^{[i]T} A^t_G X_G^{[j]} ) \otimes (X_H^{[i]T} A^t_H X_H^{[j]}) \nonumber \\
   & = \mathbf{1}^T (X_G^T A^t_G X_G ) \odot (X_H^T A^t_H X_H ) \mathbf{1}
\end{align}
}

If $H$ is a learnable hidden graph with parameters $A_H \in {\mathbb{R}^{m \times m}} $ and $X_H \in \mathbb{R}^{m \times d}$, the effective parameters are merely $X_H^T A^t_H X_H \in \mathbb{R}^{d \times d}$. That is, dimension $d$ is an important degree of freedom for learnability, which can be small for certain real-world graphs.  

% \textbf{Issue 3: Trivial solution for maximizing similarity.}  Sec.\ref{sssec:descriptive} discussed that \ckns can be optimized for maximizing total similarity score to effectively learn descriptive graph features. % without labels. 
% However, this objective combined with the RWK in Eqn.~\eqref{eq:rwgnn} leads to a trivial solution. First, notice that the hidden graph parameterization $(A_H, X_H)$ does not have restriction, which can be scaled as large as possible to maximize the number of common walks. Second, even if we restrict the values of the parameters in $[0,1]$, the hidden graphs will be learned as fully connected graphs.

%% 1. ssT works like a reweight function 
%% 2. it only reweights the final result of k-step work, without affect walks before k

%% 3. Mathematically, this equation can be transformed to the one to d x d matrix parameterization. 

%% for unsupervised training, maximize similarity directly equals to just make H fully connected. 

%\subsection{Contribution 1: 
% \vspace{-2mm}
\subsection{Color-Matching Random Walks with Efficient Computation} \label{ssec:color-matching}
%and Connections with GNNs}
% We next derive a simple but effective formulation to address . %To develop the new formulation, first 
To address \textbf{Issue 1}, we propose an improved random walk kernel \rwpp by deriving an effective formulation.
First, notice that the original RWK for labeled graphs can be rewritten using the Kronecker product $A_{G\otimes H}$ and one-hot encoded representation of the node labels. We slightly change the $G\times H$ notation by introducing a set of ``empty'' nodes $\big\{(u,v) \in V(G) \times V(H) | l(u) \neq l(v)  \big\}$ to the direct product graph $G \times H$. ``Empty'' nodes do not connect to any node hence this does not change the graph, rather they enlarge the size of $A_{G\times H}$ to be the same as $A_{G\otimes H}$. Given $X_G$ and $X_H$ as one-hot encoding of labels, the similarity matrix $S = X_H X_G^T$ is a binary valued matrix. Then, the following relation can be easily established:\looseness=-1
\begin{align}
    A_{G\times H} = \text{diag}(\mathbf{s}) A_{G\otimes H} \text{diag}(\mathbf{s}) =  \mathbf{s}\mathbf{s}^T \odot  A_{G\otimes H} \;, \label{eq:link-product}
\end{align}
where $\text{diag}(\mathbf{s})$ denotes a diagonal matrix with $\mathbf{s}$ being the diagonal. Thanks to Eqn.~\eqref{eq:link-product}, we can rewrite the original RWK in Eqn.~\eqref{eq:rw2} as:\looseness=-1
\begin{align}
   \mathcal{K}^{t}_{rw+} (G, H) = \sum_{i=1}^t \lambda_i [  \mathbf{1}^T (\mathbf{s}\mathbf{s}^T \odot  A_{G\otimes H} )^i \mathbf{1} ]\;,  \label{eq:rw-cm}
\end{align}

\noindent
which is slightly different from Eqn.~\eqref{eq:rwgnn} with $\mathbf{s}\mathbf{s}^T$ moving inside the power iteration. Although the derivation starts from labeled graphs, Eqn.~\eqref{eq:rw-cm} can be directly used for continuous attributed graphs without modification. 
Notice that this new formulation now takes all intermediary node attributes into consideration when comparing walks as intended. 
%Hence we call the modified version as RWK-CM, or color-matching RWK.

\subsubsection{Reformulation toward Efficient Computation}
The formulation in Eqn.~\eqref{eq:rw-cm} needs to compute the product graph between $G$ and $H$ which is inefficient in both memory and time. We establish an efficient computation by a property of Kronecker product, {$(A \otimes B) \text{vec}(S) = \text{vec}( BSA^T)$ \cite{van2000Kronecker}} , and rewrite the main part of Eqn.~\eqref{eq:rw-cm} step by step as follows:
{
% \footnotesize
\begin{align}
% \vspace{-2in}
 %\resizebox{0.9\hsize}{!}{
  ~ & \mathbf{1}^T (\mathbf{s}\mathbf{s}^T \odot  A_{G\otimes H} )^i \mathbf{1} \nonumber\\
  =~ & \mathbf{1}^T ( \text{diag}(\mathbf{s}) A_{G\otimes H} \text{diag}(\mathbf{s}) )^i \mathbf{1} \nonumber\\
  =~ & \mathbf{1}^T \text{diag}(\mathbf{s}^{-1})  (\text{diag}(\mathbf{s}^2) A_{G\otimes H} )^{i} \text{vec}(S) \nonumber\\
  =~ & \mathbf{1}^T \text{diag}(\mathbf{s}^{-1})  (\text{diag}(\mathbf{s}^2) A_{G\otimes H} )^{i-1} \text{diag}(\mathbf{s}^2)  \text{vec}(\textcolor{orange}{A_H} S \textcolor{orange}{A_G^T}) \nonumber \\
  =~ & \mathbf{1}^T \text{diag}(\mathbf{s}^{-1})  (\text{diag}(\mathbf{s}^2) A_{G\otimes H} )^{i-1}  \text{vec}(\textcolor{cyan}{S\odot S\odot} (\textcolor{orange}{A_H} S \textcolor{orange}{A_G^T}) ) \label{eq:rw-cm-efficient}
\end{align}
}
 
The LHS thus can be computed \textit{iteratively} by applying colored operations on the RHS repeatedly, using the procedure outlined in Algo. \ref{alg:iter} (where transpose is applied to all variables).

\begin{algorithm}[H] %[!th]
\begin{algorithmic}[1]
\caption{Fast Color-Matching RWK\hfill \COMMENT{\textcolor{gray}{and \rwppc}}}
\label{alg:iter}
   \STATE {\bfseries Input:} $G$$=$$(A_G\in \mathbb{R}^{n\times n}$, $X_G \in \mathbb{R}^{n\times d})$; $H$ $=$$(A_H \in \mathbb{R}^{m\times m}$, $X_H \in \mathbb{R}^{m\times d})$; max step $t$; \hfill \COMMENT{ \textcolor{gray}{H are parameters in \rwppc}}
   \STATE {\bfseries Init:} $Y_0$ $\leftarrow$ $X_G X_H^T$, $\;$ $Y$ $\leftarrow$ $Y_0$; \hfill\COMMENT{\textcolor{gray}{$Y_0$ $\leftarrow$ $\sigma (X_G X_H^T)$ in \rwppc}}
   \FOR{$i=1$ {\bfseries to} $t$}
   \STATE \textcolor{orange}{ $Y$ $\leftarrow A_GYA_H^T$ }
   \STATE $Y^{(i)}$ $\leftarrow $ \textcolor{cyan}{$Y_0 \odot Y$}
   \STATE $Y$ $\leftarrow $ \textcolor{cyan}{$Y_0 \odot Y^{(i)}$}
   % \STATE $Y$ $\leftarrow Y_0 \odot Y$ If $i=t$ else  \textcolor{cyan}{$Y_0 \odot Y_0 \odot Y$}
   \ENDFOR
   \STATE {\bfseries Return:} $\sum_{i,j}Y^{(t)}_{i,j} \;$ or $\;\sum_{i,j} (\sum_{l} \lambda_l \cdot Y^{(l)})_{i,j}$
\end{algorithmic}
%\vspace{-0.2in}
\end{algorithm}

\textbf{Complexity Analysis.} 
Let $G$ be the sparse input graph with $n$ nodes and $e$ edges, and let $H$ be the dense hidden graph with $m$ nodes. Eqn.~\eqref{eq:rw-cm} requires the explicit computation of Kronecker product, requiring runtime complexity $O(em^2)$ and memory complexity $O(em^2)$. In contrast, Eqn.~\eqref{eq:rw-cm-efficient} has runtime complexity $O(em + nm^2)$ and memory complexity $O(nm + m^2 + e)$.

\subsubsection{Learnable Similarity Normalization}
In Eqn.~\eqref{eq:rw-cm} we compute the random walk similarity score for each step $i$ iteratively. As each step counts the number of shared walks with length $i$, the scale of the similarity score across different steps can be considerably different, underscoring shorter walks. To combine these scores across different steps, we need to normalize the scores, which is nontrivial. To avoid hand-crafted normalization that needs hyperparameter tuning for different input, we introduce \norm that normalizes the score in a learnable way. \norm combines BatchNorm \cite{ioffe2015batch} with the sigmoid function, where BatchNorm first standardizes the score to a Normal distribution and then shifts and rescales the score with learnable parameters. Sigmoid further normalizes the score to the range $[0,1]$. We place \norm between lines 4 and 5 in Algo.~\ref{alg:iter} to normalize the score step by step, and set $\lambda_l = 1$ for all $l \in [1,t]$.\looseness=-1

%\vspace{-0.1in}
%\subsubsection{Connections with GNNs}

\subsection{Enhancing Descriptive Hidden Graphs}
\label{ssec:diversity}
Thanks to the unsupervised objective in Eqn.~\ref{eq:unsupervised} and our \rwpp, KCNs can be used to learn descriptive features.
To further enhance the descriptive ability of the hidden graphs learned by KCN, we propose \method, with two more important solutions:

\noindent\textbf{S1: Additional ``Structural Colors''.}
As discussed under \textbf{Issue 2} in Sec.~\ref{ssec:issues}, the input feature dimension $d$ (i.e. number of node attributes) is an important degree of freedom for learnability for the original RWK in \cite{nikolentzos2020random}. For RWK with color-matching, input features also play an important role as the similarity matrix $S=X_HX_G^T$ would be sparser with features that better characterize the nodes. 
Features that characterize structurally similar nodes also 
enable stronger feature matching at each step of a pair of walks between two graphs. Therefore, we propose to enrich the original node features with additional structural features in unsupervised learning of descriptive hidden graphs. As randomly initialized GNN can produce reasonable features for evaluating similarity in graph generation \cite{thompson2022evaluation}, we generate additional structural features through a fixed randomly initialized GNN to augment the original features. 

% Our assumption is that the more unique or identifiable the node features are, structurally better \hiddens we can learn through the unsupervised objective.
% better directly refer to issue 2's d 
% However, the given node features are often not unique enough to identify all the nodes.
% We thus incorporate GNN to enriching the node features with the help of structures.
% Fixed GNN with randomly initialized weights has been shown to be effective without adding extra burden of training \cite{thompson2022evaluation}.
% Structure colors can be defined as $X_{G}^{s} = \mathcal{G}(A_{G}, X_{G})$, where $\mathcal{G}$ is a fixed GNN with randomly initialized weights. 
% Then we can rewrite the objective into:

% \vspace{-0.2in}
% \begin{small}
% \begin{align}
% \begin{split}
%     \max_{H}  & \;\;\mathcal{K}^{t}_{rw}(G=(A_{G}, X_{G}), H=(A_{H}, X_{H}^{(1)})) \\
%     & + \lambda \cdot \mathcal{K}^{t}_{rw}(G=(A_{G}, X_{G}^{s}), H=(A_{H}, X_{H}^{(2)}))
% \end{split}
% \label{eqn:newobj}
% \end{align}
% \end{small}

% \vspace{-0.15in}
% \noindent
% where $\lambda$ is used for balancing.
% By separating $X_{H}$ to $X_{H}^{(1)}$ and $X_{H}^{(2)}$, we can now easily adjust the importance of learning node features and structures.

\noindent\textbf{S2: Diversity Regularization.}
When learning with more than one \hidden without any constraint, the optimization may end up learning either the most frequent or otherwise very similar patterns.
Therefore, we introduce diversity regularization $\mathcal{R}$ toward learning non-overlapping \hiddens, defined as follows:
{\small
\begin{equation}
\mathcal{R}(W_{1}, \dots, W_{k}) = \frac{2}{k(k-1)} \sum_{i=1}^{k-1}\sum_{j=i+1}^{k}{{\mathcal{K}^{t}_{rw}(W_{i}, W_{j})}}
\end{equation}
}
The overall unsupervised objective becomes maximizing the input graph to \hidden similarities, while also minimizing the pairwise RWK similarities among the \hiddens $\mathcal{R}(W_{1}, \dots, W_{k})$.

% \subsection{Practitioner's Guide}
% What to use in unsupervised and supervised settings

% E.g. diversity helps little in supervised task.

% ReLU to Sigmoid

% \input{031method-init.tex}

% We refer to the RWK with all our proposed improvements as \kername. While it can be used within various KCN architectures as in the experiments, we refer to such a model as \method w.l.o.g. 

% \vspace{-2mm}
\subsection{Connections with GNNs}
% \vspace{-1mm}
Moreover, \rwpp shares connections with Graph Convolutional Networks (GCN) \cite{kipf2017semisupervised}. If we view the hidden graph inside \rwpp as learnable parameters, line 4 of Algo.~\ref{alg:iter} is given as $Y \leftarrow A_G Y A_W^T$, which shares the \textit{same} formulation as the graph convolutional operation in GCN, ignoring the activation function. Besides convolution-like computation, \rwpp with learnable hidden graph also has a gated element-wise product as in line 5 of Algo.~\ref{alg:iter}. 
% Appendix \ref{sec:gnnconnection} elaborates on the connections for the interested reader, which we leave out of this paper's scope for brevity.
%Interestingly, one can try to incorporate additional nonlinear activation functions inside Algo. \ref{alg:iter} to transform RWK to certain neural networks, which we leave to future work. 

To demonstrate the connections with GNNs, we propose a novel GNN layer \rwppc, based on Algo.~\ref{alg:iter} (see the \textcolor{gray}{gray} part). 
The major differences between \rwppc and a normal GCNConv are:
(1) element-wise product operation with $Y_{0}$
 motivated from node color matching; and
(2) multi-step within a single convolution layer that shares the same parameter $A_{H}$ and $X_{H}$.
Additionally, we make following changes to turn it into a neural network layer:
(1) adding a sigmoid to $Y_{0}$ to normalize the scale of similarity between $0$ and $1$; and
(2) parameterizing $A_{H}$ with a fully-connected layer. 
With the learnable hidden graphs and the additional element-wise product operation, we expect \rwppc to bring better expressiveness than the GCN layer. 
We empirically demonstrate this point in Sec~\ref{ssec:gnnconnection}, across many applications.

%% file: 040exp-part1.tex
Through a series of experiments, we show that \method can be used for several unsupervised pattern mining tasks, and that each of our proposed solutions contribute to improved performance and descriptive ability.
Pattern mining, which is typically a graph algorithm subject matter, is a very difficult task to achieve via machine learning.
Since our major purpose is to demonstrate the advantages of \method over \rwk, they are evaluated on a controlled testbed with ground truth, wherein we understand the nature of the graphs.
The detailed descriptions are given in Appx.~\ref{app:detail}.

% \vspace{-2mm}
\subsection{Task 1: Simple Subgraph Matching} \label{ssec:easy}
% \vspace{-1mm}
We design two tasks where the subgraphs are easy to learn. %, to support our claims in the previous section.
The first task aims to show that \method handles \colormatch of every node pair along walks, while \rwk does not.
The second task demonstrates that \diversity regularization aids with learning non-overlapping \hiddens.
We report the matching accuracy for each experiment, 
where it is considered as a correct match when the model learns the desired subgraph pattern(s).

% \begin{observation}
% Thanks to \colormatch, \method matches the colors (labels) of every node pair in each walk, while \rwk not.
% \end{observation}

\textbf{Task 1-1.} We generate a database of $100$ bipartite graphs with heterophily, where nodes on two sides of the graph have different colors/labels {(e.g. Fig.~\ref{fig:bipartite1})}.
% It mimics the same specs (e.g., the number of graphs, edges and nodes) as a real-world dataset IMDB-BINARY.
We use one hidden graph, and the task is to learn a bipartite core; ``butterfly'' (Fig.~\ref{fig:bipartite2}), or a $3$-star with core and peripherals with different colors (Fig.~\ref{fig:bipartite3}).
% {The accuracy denotes the percentage of trials that the learned hidden graph successfully matches the correct pattern.}
Two different objectives are used; one is to maximize the total similarities from all steps, and another is to maximize the similarity only from the last step.\looseness=-1

{\footnotesize
\setlength{\tabcolsep}{3pt}
\begin{table}[t]
\caption{\emphasize{Task 1-1: Simple subgraph matching in bipartite} \emphasize{graphs.} Thanks to \colormatch, \method performs well even when the objective is based only on the last step. \label{tab:bg}}
\vspace{-3mm}
\setlength\fboxsep{0pt}
\centering{%\resizebox{0.7\columnwidth}{!}{
\begin{tabular}{c | c | c | r}
\hline
\textbf{Method} & \textbf{Objective} & \textbf{\# of Steps} & \textbf{Acc.} \\
\hline
\multirow{2}{*}{\rwk} & \multirow{2}{*}{Sum of All Steps} & $2$ & $26\%$ \\
 &  & $3$ & $100\%$ \\
\hline
\multirow{2}{*}{\rwk} & \multirow{2}{*}{Only Last Step} & $2$ & $0\%$ \\
 &  & $3$ & $100\%$ \\
\hline
\multirow{2}{*}{\method} & \multirow{2}{*}{Only Last Step} & $2$ & $100\%$ \\
 &  & $3$ & $100\%$ \\
\hline
\end{tabular}
}%}
\vspace{-2mm}
\end{table}
}
{\footnotesize
\setlength{\tabcolsep}{3pt}
\begin{table}[t]
\caption{\emphasize{Task 1-2: Simple subgraph matching in triangle} \emphasize{chains.} Diversity regularization helps \method learn non-overlapping \hiddens. \label{tab:tc}}
\vspace{-2mm}
\setlength\fboxsep{0pt}
\centering{%\resizebox{0.93\columnwidth}{!}{
\begin{tabular}{c | c | c | r r r}
\hline
\textbf{\# of Hidden Graphs} & \textbf{Method} & \textbf{Diversity} & \textbf{P1 Acc.} & \textbf{P2 Acc.} & \textbf{Both Acc.} \\
\hline
\multirow{3}{*}{$2$} & \rwk & No & $0\%$ & $0\%$ & $0\%$ \\\cline{2-6}
 & \multirow{2}{*}{\method} & No & $82\%$ & $24\%$ & $12\%$ \\
 &  & Yes & $72\%$ & $66\%$ & $44\%$ \\
\hline
\multirow{3}{*}{$3$} & \rwk & No & $0\%$ & $0\%$ & $0\%$ \\\cline{2-6}
 & \multirow{2}{*}{\method} & No & $88\%$ & $44\%$ & $32\%$ \\
 &  & Yes & $76\%$ & $80\%$ & $62\%$ \\
\hline
\multirow{3}{*}{$4$} & \rwk & No & $0\%$ & $0\%$ & $0\%$ \\\cline{2-6}
 & \multirow{2}{*}{\method} & No & $\bf 98\%$ & $68\%$ & $66\%$ \\
 &  & Yes & $84\%$ & $\bf 86\%$ & $\bf 74\%$ \\
\hline
\end{tabular}
}%}
\vspace{-3mm}
\end{table}
}

Table~\ref{tab:bg} reports the matching accuracies.
Our \method works well even if the similarity is only from the last step, regardless of the number of steps.
Since \method matches the labels of every node pair in each walk, maximizing the similarity from the last step needs to ensure the correctness of matching from previous steps at the same time.
Although \rwk works when the similarity is from all steps, it fails when the similarity is from the last step when the number of steps equals $2$.
This is because the even-step neighbors in a bipartite heterophily graph have the same color. %, and \rwk only matches the labels of the first and last node pairs in each walk.

However, this task is a special case, where the method only needs to realize that the neighbors should have the other color in the learned pattern.
That is to say, \rwk still can not solve complicated cases just by summing up the similarity from all steps.
As we will see later in this section, while \rwk always learns rudimentary patterns because of ignoring the intermediate nodes in the walks, \method learns more sophisticated ones by taking it into account.\looseness=-1

In the rest of this section, for fair comparison, we use ``Sum of All Steps'' as the objective for \rwk, and ``Only Last Step'' for \method, which performs well and simplifies the optimization.

\textbf{Task 1-2.} To test diversity regularization, we generate a database with $100$ node-labeled triangle chains, containing two frequent patterns (e.g. Fig.~\ref{fig:chain1}).
Each triangle is either pattern P1 ({Fig.~\ref{fig:chain2}}) with probability $60\%$ or otherwise P2 ({Fig.~\ref{fig:chain3}}) with lower frequency.
% The parameters are randomly initialized for $100$ times to repeat the experiments.
%Accuracy for both denotes both P1 and P2 are included in the learned \hiddens.
The number of steps is set to $3$, which is efficient and sufficient to capture both homophily (1-step) and heterophily (2-step) neighbors.

Table~\ref{tab:tc} reports the results, where accuracy depicts if \textit{both} P1 and P2 are learned by the \hiddens.
Even without {diversity regularization}, \method learns the most frequent pattern P1 with high accuracy.
When diversity regularization is applied, accuracies for the second frequent pattern P2 and both patterns increase.
The increase is larger when \method is trained more flexibly with a larger number of hidden graphs to be learned. 
Notably, RWNN with vanilla RWK fails to learn either of the patterns.
As it prefers the more frequent colors, it often learns all the node colors to be the same.\looseness=-1
%; i.e. red (most frequent), or purple (in smaller subset).

\begin{figure}[t]
\centering
\captionsetup[subfigure]{width=0.8in}
\subfloat[Tailed Triangle: Ground Truth]{\includegraphics[scale=0.4]{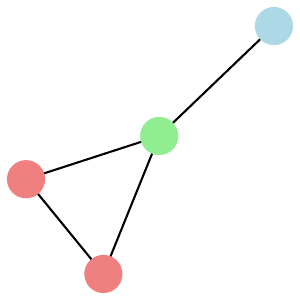}\label{fig:a1}}
\hspace{6mm}
\subfloat[Hidden Graph by \method]{\includegraphics[scale=0.4]{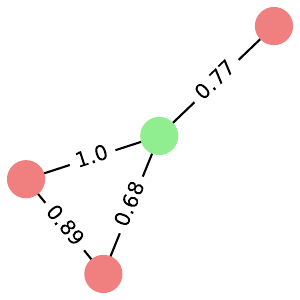}\label{fig:a2}} 
\hspace{6mm}
\subfloat[Hidden Graph by \rwk]{\includegraphics[scale=0.4]{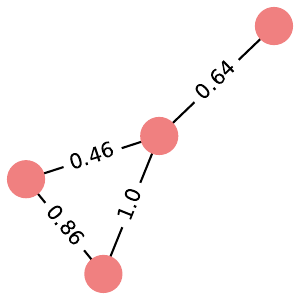}\label{fig:a3}} \\
\vspace{-2mm}
\captionsetup[subfigure]{width=2in}
\subfloat[Table of results. Lower GED is better.]{
% \resizebox{0.74\columnwidth}{!}{
{\footnotesize
\setlength{\tabcolsep}{3pt}
\begin{tabular}{c | c | C{1.8cm} c}
\hline
\textbf{Method} & \textbf{\# of Steps} & \textbf{GED \textcolor{blue}{w/} Node Labels} & \textbf{$p$-value} \\
\hline
\rwk    & $2$ & $3.35 \pm 0.41$ & $3.1 e\text{-} 05^{***}$ \\
\rwk    & $4$ & $3.15 \pm 0.34$ & $3.2 e\text{-} 03^{**}$ \\
\rwk    & $6$ & $3.25 \pm 0.39$ & $4.7 e\text{-} 04^{***}$ \\
\hline
\method & $2$ & $2.87 \pm 0.58$ & $0.20$ \\
\method & $4$ & $2.82 \pm 0.64$ & $0.33$ \\
\hline
\method & $6$ & $\bf 2.76 \pm 0.86$ & - \\
\hline
\end{tabular}\label{tab:ttri}
}%}
}
\vspace{-3mm}
\caption{\emphasize{Task 2-1: GED-based evaluation on tail-triangles.}} \label{fig:ttri}
\vspace{-7mm}
\end{figure}

\begin{figure}[t]
\centering
\captionsetup[subfigure]{width=0.8in}
\subfloat[Ring:\\Ground Truth]{\includegraphics[scale=0.4]{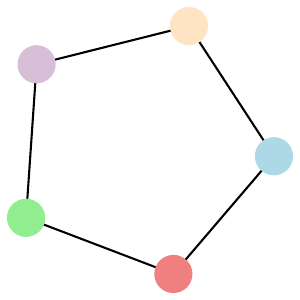}\label{fig:r1}}
\hspace{6mm}
\subfloat[Hidden Graph by \method]{\includegraphics[scale=0.4]{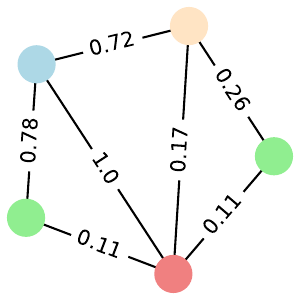}\label{fig:r2}} 
\hspace{6mm}
\subfloat[Hidden Graph by \rwk]{\includegraphics[scale=0.4]{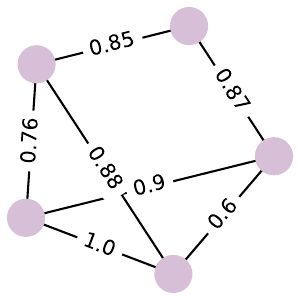}\label{fig:r3}} \\
\vspace{-2mm}
\captionsetup[subfigure]{width=2in}
\subfloat[Table of results. Lower GED is better.]{
% \resizebox{0.74\columnwidth}{!}{
{\footnotesize
\setlength{\tabcolsep}{3pt}
\begin{tabular}{c | c | C{1.8cm} c}
\hline
\textbf{Method} & \textbf{\# of Steps} & \textbf{GED \textcolor{blue}{w/} Node Labels} & \textbf{$p$-value} \\
\hline
\rwk    & $2$ & $7.08 \pm 0.64$ & $8.4 e\text{-} 14^{***}$ \\
\rwk    & $4$ & $7.16 \pm 0.58$ & $2.7 e\text{-} 14^{***}$ \\
\rwk    & $6$ & $6.92 \pm 0.75$ & $2.1 e\text{-} 11^{***}$ \\
\hline
\method & $2$ & $5.58 \pm 0.65$ & $0.21$ \\
\method & $4$ & $5.59 \pm 0.73$ & $0.21$ \\
\hline
\method & $6$ & $\bf 5.46 \pm 0.86$ & - \\
\hline
\end{tabular}\label{tab:ring}
}%}
}
\vspace{-3mm}
\caption{\emphasize{Task 2-1: GED-based evaluation on rings.}} \label{fig:ring}
\vspace{-3mm}
\end{figure}

\subsection{Task 2: GED-Based Evaluation} \label{ssec:hard}
To further show the advantages of \method, we design two more tasks each with two different testbeds.
For evaluation, these experiments consider a database containing $100$ {identical} graphs, which is used as the ground truth, i.e. only one hidden graph is used in both tasks.
As the ground truth is more complex than the ones in Task 1, and it is difficult to learn the exact graph, we use graph edit distance (GED) \cite{sanfeliu1983distance} to measure how close the learned \hidden is to the ground truth (the lower the better).
While GED with node labels induces a penalty for editing the labels, GED without node labels purely focuses on the graph structure.
The first task studies labeled graphs with different number of steps, and shows that \method outperforms \rwk thanks to \colormatch.
The second task demonstrates the effectiveness of adding ``\structure'', which improves both the learned structure and labels.
%For each of the tasks, we use the same initialization set to run the experiment for several times.
We report $p$-values based on the paired $t$-test that quantify differences between two GED values statistically.\looseness=-1

% \begin{observation}
% \method learns better \hiddens than \rwk in GED with node labels on node-labeled graphs.
% \end{observation}

\textbf{Task 2-1.} We design two testbeds using node-labeled tailed triangles and rings, as shown in Fig.~\ref{fig:a1} and \ref{fig:r1}, respectively (best in color).
We learn the \hidden with the same number of nodes as the ground truth graph. 
% The parameters are randomly initialized for $50$ times.
Tables~\ref{tab:ttri} and \ref{tab:ring} report the GED comparison.\looseness=-1
%The \hiddens learned by 

\method achieves consistently lower GED than %the ones learned by 
\rwk, demonstrating the importance of incorporating \colormatch into the RWK.
Experiments on both testbeds show that there is no clear choice for the number of steps, i.e., higher is not always better, where the p-values are high within \method.
We visualize the learned \hiddens by removing the edges with the smallest edge weights.
Fig.~\ref{fig:a2} shows that \method successfully assigns the green node with degree $3$ in the correct position.
Since the blue node only has degree $1$, \method reasonably learns to maximize the objective by adding one more red node in the \hidden, which is the most frequent color;
in Fig.~\ref{fig:a3}, \rwk fails to handle the intermediate nodes, and hence includes only the most frequent color in the learned \hidden.
We observe a similar behavior in Fig.~\ref{fig:r2} and \ref{fig:r3}.
While \method pays much attention to learning the correct node labels, \rwk gives a rudimentary result, where all the nodes have the same labels.

{\footnotesize
\setlength{\tabcolsep}{3pt}
\begin{table}[t]
\caption{\emphasize{Task 2-2: GED-based evaluation on $3$-regular unla-} \emphasize{beled graph.} Being used as unique identifiers of nodes, \structure are shown to be as effective as identity matrix. \label{tab:3reg}}
\vspace{-2mm}
\setlength\fboxsep{0pt}
\centering{%\resizebox{0.85\columnwidth}{!}{
\begin{tabular}{c | C{2cm} C{2cm} c}
\hline
\textbf{Method} & \textbf{Additional Features} & \textbf{GED \textcolor{red}{w/o} Node Labels} & \textbf{$p$-value} \\
\hline
\rwk    & None     & $4.38 \pm 0.65$ & - \\
\hline
\rwk    & Identity & $4.41 \pm 0.56$ & $0.57$ \\
\method & Identity & $\bf 3.89 \pm 0.48$ & $4.4 e\text{-} 05^{***}$ \\
\hline
\rwk    & SC       & $4.45 \pm 0.70$ & $0.72$ \\
\method & SC       & $\bf 4.10 \pm 0.50$ & $0.010^{*}$ \\
\hline
\end{tabular}
}%}
\vspace{-2mm}
\end{table}
}

{\footnotesize
\setlength{\tabcolsep}{3pt}
\begin{table}[t]
\caption{\emphasize{Task 2-2: GED-based evaluation on $2$-regular labeled} \emphasize{graph.} Both \colormatch and \structure improve the quality of structure and label learned by \hidden. \label{tab:2reg}}
\vspace{-2mm}
\setlength\fboxsep{0pt}
\centering{%\resizebox{0.9\columnwidth}{!}{
\begin{tabular}{c | C{1.6cm} C{1.8cm} C{1.3cm} C{1.3cm}}
\hline
\textbf{Method} & \textbf{Additional Features} & \textbf{GED \textcolor{red}{w/o} Node Labels} & \textbf{$p$-value w/ Row 1} & \textbf{$p$-value w/ Row 2} \\
\hline
\rwk    & None     & $5.25 \pm 0.64$ & - & - \\
\hline
\method & None     & $5.02 \pm 0.63$ & $0.049^{*}$ & - \\
\hline
\method & Identity & $\bf 4.81 \pm 0.70$ & $1.0 e\text{-} 03^{**}$ & $0.043^{*}$ \\
\method & SC       & $\bf 4.82 \pm 0.70$ & $1.1 e\text{-} 03^{**}$ & $0.043^{*}$ \\
\hline
\end{tabular}
}%}
\centering{%\resizebox{0.9\columnwidth}{!}{
\begin{tabular}{c | C{1.6cm} C{1.8cm} C{1.3cm} C{1.3cm}}
\hline
\textbf{Method} & \textbf{Additional Features} & \textbf{GED \textcolor{blue}{w/} Node Labels} & \textbf{$p$-value w/ Row 1} & \textbf{$p$-value w/ Row 2} \\
\hline
\rwk    & None     & $7.25 \pm 0.64$ & - & - \\
\hline
\method & None     & $6.60 \pm 0.93$ & $1.8 e\text{-} 04^{***}$ & - \\
\hline
\method & Identity & $\bf 6.12 \pm 1.01$ & $2.9 e\text{-} 08^{***}$ & $1.8 e\text{-} 03^{**}$ \\
\method & SC       & $\bf 6.25 \pm 1.01$ & $7.2 e\text{-} 07^{***}$ & $0.015^{*}$ \\
\hline
\end{tabular}
}%}
\vspace{-3mm}
\end{table}
}

{\footnotesize
\setlength{\tabcolsep}{3pt}
\begin{table*}[t]
\caption{{\emphasize{Graph anomaly detection} on $10$ real-world datasets. Recall is reported. iGAD using our proposed \rwpp as structural feature extractor outperforms original iGAD on all datasets.} \label{tab:igad}}
\setlength\fboxsep{0pt}
\centering{%\resizebox{0.9\linewidth}{!}{
\begin{tabular}{l | cccccccccc}
\hline
\textbf{Dataset} & \textbf{MCF-7} & \textbf{MOLT-4} & \textbf{PC-3} & \textbf{SW-620} & \textbf{NCI-H23} & \textbf{OVCAR-8} & \textbf{P388} & \textbf{SF-295} & \textbf{SN12C} & \textbf{UACC-257} \\
\hline
% \# of Graphs & $27,770$ & $39,765$ & $27,509$ & $40,532$ & $40,353$ & $40,516$ & $41,472$ & $40,271$ & $40,004$ & $39,988$ \\
% \# of Anomalies & $2,294$ & $3,140$ & $1,568$ & $2,410$ & $2,057$ & $2,079$ & $2,298$ & $2,025$ & $1,955$ & $1,643$ \\
% Avg. \# of Nodes & $26.4$ & $26.1$ & $26.4$ & $26.1$ & $26.1$ & $26.1$ & $22.1$ & $26.1$ & $26.1$ & $26.1$ \\
% Avg. \# of Edges & $28.5$ & $28.1$ & $28.5$ & $28.1$ & $28.1$ & $28.1$ & $23.6$ & $28.1$ & $28.1$ & $28.1$ \\
% \hline
iGAD + \rwori & 75.1$\pm$1.1 & 74.1$\pm$0.8 & 77.9$\pm$1.2 & 78.6$\pm$0.9 & 78.7$\pm$1.3 & 78.8$\pm$0.3 & 83.1$\pm$1.7 & 78.3$\pm$1.1 & 79.4$\pm$0.6 & 78.0$\pm$1.0 \\
iGAD + \rwpp & \textbf{76.4$\pm$0.6} & \textbf{74.3$\pm$1.0} & \textbf{78.8$\pm$1.1} & \textbf{79.2$\pm$0.5} & \textbf{79.5$\pm$2.2} & \textbf{79.2$\pm$0.9} & \textbf{84.0$\pm$1.4} & \textbf{78.5$\pm$0.9} & \textbf{79.5$\pm$1.6} &  \textbf{79.5$\pm$0.7} \\
\hline
\end{tabular}
}%}
\end{table*}
}

{\footnotesize
\setlength{\tabcolsep}{3pt}
\begin{table*}[t]
\caption{\emphasize{Graph classification} on $10$ real-world datasets. Accuracy is reported. Although \rwpp is built to capture descriptive features, it is competitive on most datasets. \label{tab:cls}}
\setlength\fboxsep{0pt}
\centering{%\resizebox{0.9\linewidth}{!}{
\begin{tabular}{l | cccccccccc}
\hline
\textbf{Dataset} & \textbf{MUTAG} & \textbf{D\&D} & \textbf{NCI1} & \textbf{PROTEINS} & \textbf{MUTAGEN} & \textbf{TOX21} & \textbf{ENZYMES} & \textbf{IMBD-B} & \textbf{IMDB-M} & \textbf{REDDIT} \\
\hline
% \# of Graphs & $188$ & $1,178$ & $4,110$ & $1,113$ & $4,337$ & $8,169$ & $600$ & $1,000$ & $1,500$ & $10,155$ \\
% \# of Classes & $2$ & $2$ & $2$ & $2$ & $2$ & $2$ & $6$ & $2$ & $3$ & $2$ \\
% Avg. \# of Nodes & $17.9$ & $284.3$ & $29.9$ & $39.1$ & $30.3$ & $18.1$ & $32.6$ & $19.8$ & $13.0$ & $23.9$ \\
% Avg. \# of Edges & $19.8$ & $715.7$ & $32.3$ & $72.8$ & $30.8$ & $18.5$ & $62.1$ & $96.5$ & $65.9$ & $25.0$ \\
% \hline
KerGNN + \rwori & 81.9$\pm$5.3 & \textbf{75.2$\pm$1.5} & 71.6$\pm$2.6 & 75.3$\pm$1.2 & 74.4$\pm$2.4 & 89.1$\pm$0.3 & \textbf{47.3$\pm$3.9} & 71.2$\pm$2.1 & 48.1$\pm$2.9 & 77.2$\pm$0.5 \\
KerGNN + \rwpp & \textbf{83.0$\pm$6.4} & 74.8$\pm$2.4 & \textbf{72.3$\pm$1.3} & \textbf{76.2$\pm$1.2} & \textbf{75.1$\pm$1.0} & \textbf{89.2$\pm$0.3} & 44.0$\pm$2.7 & \textbf{71.6$\pm$1.0} & \textbf{49.2$\pm$0.6} & \textbf{77.5$\pm$0.6} \\
\hline
\end{tabular}
}%}
\end{table*}
}

\textbf{Task 2-2.} 
Two more testbeds are designed to evaluate \structure. %, and run the experiments for $20$ times.
The number of steps is set to $3$, which is effective and efficient.
The first database contains $3$-regular unlabeled graphs (Fig.~\ref{fig:h1}).
We evaluate \method and \rwk by GED \textit{without} node labels, focusing on the quality of the learned structure.
Our assumption is, if the structural identifiers (node labels) are more unique, then the \hidden can learn better graph structure.
Therefore, we assume identity matrix as the best features in the evaluation, though it is not generalizable to the real datasets.
We create the \structure by 
% feeding an identity matrix into 
a fixed and randomized Graph Attention Networks (GAT) \cite{velivckovic2017graph}.
The results are reported in Table~\ref{tab:3reg}.
Our proposed \method using identity matrix as features receives the lowest GED without node labels, as expected.
\method using \structure has competitive GED compared with using identity matrix, while being more generalizable.
These results also empirically prove our assumption that using more unique identifiers as node features helps the \hidden to learn better structure.
In contrast, \rwk fails to utilize the features even if they are extremely informative.\looseness=-1

In the second testbed, we study a database containing $2$-regular node-labeled graphs (i.e. $6$-ring, Fig.~\ref{fig:h2}), and report the results in Table~\ref{tab:2reg}.
Only by replacing \rwk with our proposed \method, the quality of learned \hidden improves not only on labels, but also on the structure.
In addition to the node labels, we further incorporate \structure into \method, and find the learned \hidden improves even better, demonstrating the effectiveness of \structure.
Notably, using identity matrix as features results in only slightly lower GED than using \structure.\looseness=-1

%% file: 050exp-part2.tex
In this section, the experiments is composed of two parts.
In the first part, we demonstrate that different KCN architectures can perform better by employing our proposed \rwpp.
In the second part, we compare our proposed \rwppc with GCNConv, and empirically show that it has better expressiveness.
% Datasets details are provided in Appendix~\ref{app:data}.
% For graph anomaly detection and classification, we perform $5$-fold cross-validation and split $10\%$ of training set as the validation set.
The details of dataset statistics and hyperparameter search are in Appx.~\ref{app:rep}.\looseness=-1

\subsection{\rwpp: Employed to Different Architectures} \label{ssec:rwppexp}
We conduct three graph learning tasks for evaluating \kername.
Since we focus on improving \rwori across many tasks, rather than outperforming task-specific state-of-the-art methods, the experiments concentrate on comparing models using RWK versus \kername.

\subsubsection{\textbf{iGAD on Graph Anomaly Detection}}~

\noindent\textbf{Datasets.}
% We evaluate our \rwpp on supervised imbalanced graph anomaly detection on $10$ real-world datasets publicly available from PubChem \cite{yan2008mining}.
We evaluate \rwpp on supervised graph anomaly detection with $10$ real-world datasets from PubChem \cite{yan2008mining}, as in \cite{zhang2022dual}.
Each graph is a chemical compound and labeled by its outcome from anti-cancer screen tests (active or inactive).
The classes are highly imbalanced, where the ratio of the active samples is at most $12\%$, which are treated as the anomalous cases.
We perform $5$-fold cross-validation and split $10\%$ of training set as the validation set.\looseness=-1
% We report the statistics in Table~\ref{tab:statad}.

\noindent\textbf{Settings.}
iGAD \cite{zhang2022dual} incorporates \rwori as a structural feature extractor to identify graph-level anomalies. 
For comparison we replace it with \rwpp using \norm, with one-hot node degrees as the node features.
Recall is used for both evaluation and model selection, as in the iGAD paper.
%of recall
%Recall is used for measuring the performance and selecting the best model based on the validation set.
% We perform stratified $5$-fold cross-validation and split $10\%$ of data from the training set as the validation set.
% Hyperparameter search based on the validation set is done for each fold (see Appendix~\ref{app:hyper}).

\noindent\textbf{Results.} 
We report the average performance and standard deviation (stdev) in Table~\ref{tab:igad}.
iGAD with our proposed \rwpp outperforms the original model on \textit{all} datasets ($p$-val $<$$0.001$).
This suggests that the \hiddens learned through \rwpp are consistently better than the ones extracted by \rwori, assisting iGAD in better pointing out the anomalous graphs that deviate from these patterns.

\subsubsection{\textbf{KerGNN on Substructure Counting}}~

\noindent\textbf{Datasets.}
We evaluate \rwpp on substructure counting with a simulated dataset from \cite{chen2020can}, following the same setting in \cite{zhao2022from}, and the task is to predict the normalized count of substructures.
This dataset includes four tasks, and the evaluation for different tasks are run separately.
The dataset provides the training, validation, and testing sets with $1,500/1,000/2,500$ graphs, respectively. 
One-hot node degrees are used as the node features.
% This experiment aims to show that \rwpp extracts better features than \rwori; however, the performance of worse features can be ameliorated by using a more complex model. 
% Thus, we constrain the complexity of the model to be low by using only one message-passing layer.
% We report the statistics in Table~\ref{tab:statsgc}.

\noindent\textbf{Settings.}
KerGNN \cite{feng2022kergnns} uses \rwori to compare the similarity between the learnable \hiddens and the egonets of nodes in a graph.
The similarity from different learnable \hiddens are used as the features for message passing.
In this experiment, we replace \rwori inside KerGNN with \rwpp.
Mean absolute error (MAE) is used to measure the accuracy of counts.

% We constrain the model to be very simple, in order to reflect the true quality of extracted features.
% Hyperparameter search based on the validation set is done (see Appendix~\ref{app:hyper}).
% We report the Mean absolute error (MAE) from the test set.

\noindent\textbf{Results.}
As shown in Table~\ref{tab:sgc},
KerGNN with \rwpp outperforms KerGNN with \rwori in $3$ out of $4$ tasks.
\norm is shown to effectively improve the performance of both methods by normalizing the similarity in each step.
Without it, the similarity explodes after a number of steps, and is always dominated by the latest step.
Although \rwori performs better in counting stars, there are few paths to walk within a star, which decreases the necessity of adopting a kernel that is more accurate on similarity.

{\footnotesize
\setlength{\tabcolsep}{3pt}
\begin{table}[t]
\caption{{\emphasize{Substructure counting} on a simulated dataset. MAE is reported. \rwpp wins in $3$ out of $4$ tasks, and \norm is shown to be effective.} \label{tab:sgc}}
\setlength\fboxsep{0pt}
\centering{%\resizebox{1\columnwidth}{!}{
\begin{tabular}{l | cccc}
\hline
\textbf{Task} & \textbf{Triangle} & \textbf{Tailed Tri.} & \textbf{Star} & \textbf{4-Cycle} \\
\hline
KerGNN + \rwori & 0.1170 & 0.1346 & 0.1333 & 0.2153 \\
KerGNN + \rwori + \norm & 0.1065 & 0.1251 & \textbf{0.0999} & 0.2140 \\
\hline
KerGNN + \rwpp & 0.1206 & 0.1246 & 0.1750 & 0.2078 \\
KerGNN + \rwpp + \norm & \textbf{0.0802} & \textbf{0.1240} & 0.1312 & \textbf{0.1884} \\
\hline
\end{tabular}
}%}
\end{table}
}

\begin{figure}[t]
\centering
% \captionsetup[subfigure]{width=0.85in}
\subfloat{\includegraphics[scale=0.42]{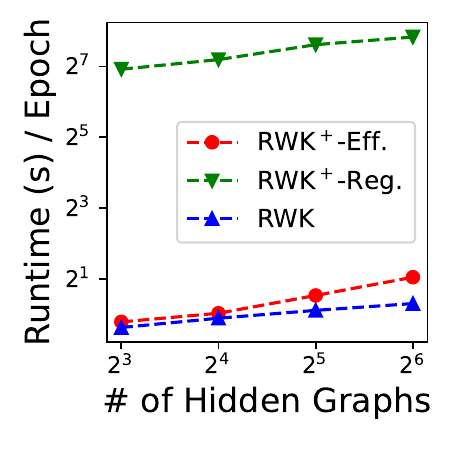}\label{fig:scale1}}
\hspace{3mm}
\subfloat{\includegraphics[scale=0.42]{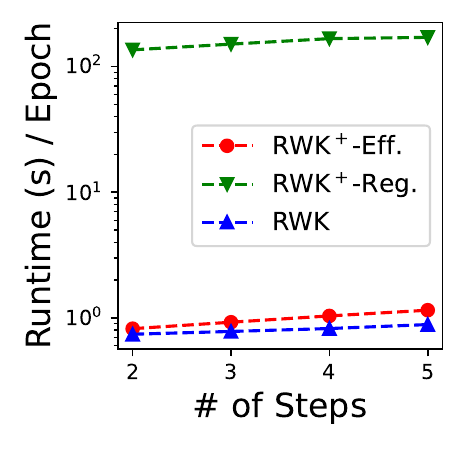}\label{fig:scale2}}
\vspace{-4mm}
\caption{\emphasize{Runtime of KerGNN+RWK$^{+}$} computed by regular Eqn. \eqref{eq:rw-cm} vs. efficient Eqn. \eqref{eq:rw-cm-efficient}, also compared to vanilla \rwori.}
\label{fig:scale}
\end{figure}

\subsubsection{\textbf{KerGNN on Graph Classification}}~

\noindent\textbf{Datasets.}
% We employ \rwpp for graph classification on $10$ real-world TUDatasets \cite{Morris2020}.
We evaluate \rwpp on graph classification with $10$ real-world datasets from TUDataset \cite{Morris2020}, as in \cite{feng2022kergnns}.
We use the node labels given by bio-informatics datasets (first $7$), and the one-hot node degrees for the social interaction datasets (last $3$).
We perform $5$-fold cross-validation and split $10\%$ of training set as the validation set.\looseness=-1
% We report the statistics in Table~\ref{tab:statcls}.

\noindent\textbf{Settings.}
Similar to substructure counting, KerGNN is used with \rwori versus \rwpp.
For fair comparison, \norm is employed for both models.
% We perform stratified $5$-fold cross-validation and split $10\%$ of data from the training set as the validation set.
% Hyperparameter search based on the validation set is done for each fold (see Appendix~\ref{app:hyper}).
We report average accuracy and stdev.

\noindent\textbf{Results.}
Table~\ref{tab:cls} shows that
although \rwpp is designed with descriptive, structural graph features in mind, 
it offers competitive performance on most classification tasks ($p$-val $<$$0.1$).

\noindent\textbf{Scalability.}
Finally, we verify \rwpp's scalability empirically, varying (1) the number of \hiddens and (2) the number of steps on the NCI1 dataset, and report the runtime per epoch during training.
In (1), the number of step is set to be $2$, and in (2), the number of hidden graphs is set to be $8$.
As shown in Fig.~\ref{fig:scale}, KerGNN with \rwpp is slightly slower than with \rwori, although the overhead is negligible ($<1$ sec.), and scales \textit{linearly}, with significant performance gains over regular color-matching computation.\looseness=-1

\subsection{\rwppc: Connections with GNNs} \label{ssec:gnnconnection}
To show that \rwppc is more expressive than GCNConv, a GCN message-passing layer \cite{kipf2017semisupervised}, we compare them across both node- and graph-level tasks.
In all experiments, we rigorously ensure that both kinds of layers share exactly the same message-passing backbone.\looseness=-1

\subsubsection{\textbf{Node Classification}}~

{\footnotesize
\setlength{\tabcolsep}{3pt}
\begin{table}[t]
\caption{\emphasize{Node classification} on $6$ real-world datasets. \rwppc wins in most tasks on accuracy. \label{tab:gnn}}
\vspace{-2mm}
\setlength\fboxsep{0pt}
\centering{%\resizebox{1\columnwidth}{!}{
\begin{tabular}{l | cccccc}
\hline
\textbf{Dataset} & \textbf{Cora} & \textbf{CiteSeer} & \textbf{PubMed} & \textbf{Cham.} & \textbf{Squirrel} & \textbf{Actor} \\
% \hline
% \# of Nodes & $2,709$ & $3,327$ & $19,717$ & $2,277$ & $5,201$ & $29,926$ \\
% \# of Edges & $5,429$ & $4,732$ & $44,338$ & $36,101$ & $216,933$ & $7,600$ \\
% \# of Classes & $7$ & $6$ & $3$ & $5$ & $5$ & $5$ \\
\hline
GCNConv & 85.8$\pm$0.7 & 73.4$\pm$0.5 & \textbf{88.0$\pm$0.2} & \textbf{69.7$\pm$0.9} & \textbf{55.7$\pm$0.4} & 28.3$\pm$0.6 \\
\rwppc & \textbf{88.3$\pm$0.5} & \textbf{76.7$\pm$0.2} & \textbf{88.1$\pm$0.2} & \textbf{69.3$\pm$1.9} & 49.7$\pm$1.3 & \textbf{36.0$\pm$0.2} \\
\hline
\end{tabular}
}%}
\end{table}
}

\noindent\textbf{Datasets.}
We evaluate \rwppc on node classification with $6$ datasets, including homophily graphs (first $3$) \cite{yang2016revisiting} and heterophily graphs (last $3$) \cite{rozemberczki2021multi,Pei2020Geom-GCN}.
Each dataset is split into $60\%/20\%/20\%$ for training, validation, and testing, respectively.

\noindent\textbf{Results.}
In Table~\ref{tab:gnn}, although expressiveness does not necessarily play a key role in achieving better accuracy on node-level tasks, \rwppc still has competitive or better performance than GCNConv in most datasets. 
We also report the run time per epoch on the largest dataset PubMed in Table~\ref{tab:gnntime}, where \rwppc only creates negligible computational overhead (less than $0.05$ second).

\subsubsection{\textbf{Twitter Bot Detection}}~

\noindent\textbf{Datasets.}
We evaluate \rwppc on a web application, namely bot detection in the TwiBot-22 dataset \cite{feng2022twibot}.
This dataset contains a web-scale Twitter social network with one million users, where $86\%$ of them are human, and the rest $14\%$ are bots.
We keep only the edges with types ``followed'' and ``following'', and make it undirected.
The node features are embeddings of user descriptions, transformed by BERT \cite{devlin2018bert}.
The dataset provides the training, validation, and testing sets with $70\%/20\%/10\%$ nodes, respectively.

\noindent\textbf{Results.}
In Table~\ref{tab:botdetect}, we find that \rwppc outperforms GCNConv on F1-score.
This suggests that \rwppc has better ability to detect the bots by better utilizing the graph structure.

{\footnotesize
\setlength{\tabcolsep}{3pt}
\begin{table}[t]
\caption{\emphasize{Runtime of RWK$^+$Conv}, with negligible overhead. \label{tab:gnntime}}
\vspace{-1mm}
\setlength\fboxsep{0pt}
\centering{%\resizebox{0.7\columnwidth}{!}{
\begin{tabular}{l | cccc}
\hline
\textbf{Step Length} & \textbf{2} & \textbf{3} & \textbf{4} & \textbf{5} \\
\hline
GCNConv & 0.0269 & - & - & - \\
\rwppc & 0.0371 & 0.0464 & 0.0522 & 0.0619 \\
\hline
\end{tabular}
}%}
\end{table}
}

{\footnotesize
\setlength{\tabcolsep}{3pt}
\begin{table}[t]
\caption{\emphasize{Twitter bot detection} on a real-world web-scale dataset. \rwppc wins on F1-score. \label{tab:botdetect}}
\vspace{-1mm}
\setlength\fboxsep{0pt}
\centering{%\resizebox{0.4\columnwidth}{!}{
\begin{tabular}{l | cccc}
\hline
\textbf{Dataset} & \textbf{TwiBot-22} \\
\hline
GCNConv & 53.7$\pm$0.2 \\
\rwppc & \textbf{55.0$\pm$0.2} \\
\hline
\end{tabular}
}%}
\end{table}
}

\newpage
\subsubsection{\textbf{Graph Regression and Classification}}~

\noindent\textbf{Datasets.}
We evaluate \rwppc on graph regression and classification with three real-world datasets, ZINC \cite{dwivedi2020benchmarking}, ogbg-molhiv and ogbg-molpcba \cite{hu2020open}. 
% To demonstrate that the \rwppc is more expressive, we compare it with the GCN on three datasets.
Note that \method does not use edge features.\looseness=-1

\noindent\textbf{Results.}
We include an additional baseline GINConv from GIN \cite{xu2018how}.
In Table~\ref{tab:gnnogb}, we find that \rwppc outperforms both baselines significantly across all datasets and tasks.
This empirically demonstrates the better expressiveness of \rwppc than GCNConv.\looseness=-1

\subsubsection{\textbf{Summary and Future Work}}~

\noindent
All results strongly suggest the better expressiveness of our proposed \rwppc, especially on graph-level tasks, and its connection to GCN motivates novel convolutional layers for better model design.
This offers a direction with large potential to investigate further in the future. 
We also want to point out that the current design of the \rwppc does not take edge features into consideration. 
Extending it to handle edge features by matching edge colors at every step of random walk could be a potential future work.
% This experiment empirically demonstrates the connections with GNNs, and points out a potential framework that can be investigated in the future.

{\footnotesize
\setlength{\tabcolsep}{3pt}
\begin{table}[t]
\caption{\emphasize{Graph regression and classification} on $3$ real-world datasets. \rwppc wins in all tasks. \label{tab:gnnogb}}
\vspace{-1mm}
\setlength\fboxsep{0pt}
\centering{%\resizebox{1\columnwidth}{!}{
\begin{tabular}{l | cccccc}
\hline
\textbf{Dataset} & \textbf{ZINC} & \textbf{ogbg-molhiv} & \textbf{ogbg-molpcba} & \\
\hline
Metric & MAE~$\downarrow$ & ROC-AUC~$\uparrow$ & AP~$\uparrow$ \\
\hline
GCNConv & 0.3258$\pm$0.0067 & 76.06$\pm$0.97 & 20.20$\pm$0.24 \\
GINConv & 0.2429$\pm$0.0033 & 77.78$\pm$1.30 & 22.66$\pm$0.28 \\
\rwppc & \textbf{0.2082$\pm$0.0025} & \textbf{78.61$\pm$0.61} & \textbf{24.90$\pm$0.12}  \\
\hline
\end{tabular}
}%}
\end{table}
}

%% file: 060conclusion.tex
%%%%%%%%%%%%%%%%%%%%%%%%%%%%%%%%%%%%%%%
% AUTHOR: Christos Faloutsos
% INSTITUTION: CMU
% DATE: April 2019
% GOAL: to streamline the paper presentations
%%%%%%%%%%%%%%%%%%%%%%%%%%%%%%%%%%%%%%%
In this paper, we first presented \kername, an improved random walk kernel with end-to-end learnable hidden graphs that can be used by various KCNs. 
%, and demonstrated its mathematical connection with GNNs, using which we proposed \norm for learnable similarity normalization. We further improved \method in the unsupervised setting by enriching node labels with \structure and diversity regularization.
\kername incorporates color-matching along the walks that we showed can be efficiently computed in iterations, and combines similarities across steps in a learnable fashion.
We then proposed \method, a KCN that learns descriptive hidden graphs with an unsupervised objective and \kername.
Thanks to additional ``structural colors'' and diversity regularization, it learns hidden graphs that better reflect the frequent and distinct graph patterns.
Moreover, based on the mathematical connection of \kername with GNNs, we propose a novel GNN layer \rwppc, that extracts expressive graph representations.
Experiments showed \kername's descriptive learning ability on various unsupervised graph pattern mining tasks, as well as its advantages when employed within various KCN architectures on several supervised graph learning tasks.
Furthermore, we showed that our proposed \rwppc layer outperforms GCN, especially in the graph-level tasks by a large margin.\looseness=-1
% All source code and datasets are available (anonymously) at \codeurl. 

%% file: 070appendix.tex
\section{Unsupervised Pattern Mining} \label{app:detail}
As the optimization is prone to local minima, we randomly initialize the parameters several times and report the average result.
The random initialization is done on the parameters of hidden graph(s) and its/their node features.
The parameters of hidden graphs are uniformly initialized between $-1$ to $1$. 
The parameters of the node features of the hidden graphs are uniformly initialized between $0$ to $1$.
Moreover, we have two differences from \rwk.
First, \rwk uses the ReLU function right after constructing the \hiddens, which makes the gradients of edges, whose weights are initialized with negative values, become zero.
We thus replace the ReLU function with Sigmoid function to properly learn the \hiddens. 
Second, in order to prevent the model stuck at local minimums, we use SGD with a large momentum as the optimizer.

% c) doing degree-normalization on both adjacency matrices.

\subsection{Task 1: Simple Subgraph Matching}
\begin{figure}[h]
\centering
\subfloat[Bipartite]{\includegraphics[scale=0.4]{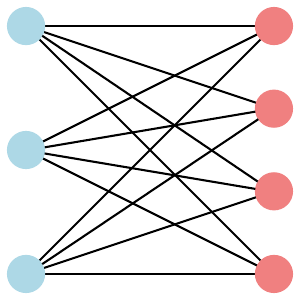}\label{fig:bipartite1}}
\hspace{2mm}
\subfloat[Butterfly]{\includegraphics[scale=0.4]{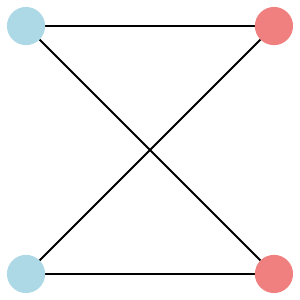}\label{fig:bipartite2}} 
\hspace{2mm}
\subfloat[Star]{\includegraphics[scale=0.4]{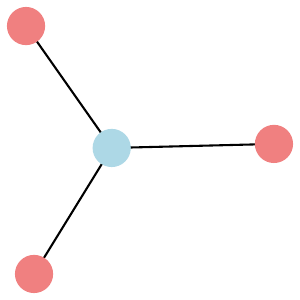}\label{fig:bipartite3}}
\vspace{-2mm}
\caption{Simple Subgraph Matching in Bipartite Graphs} \label{fig:bipartite}
\vspace{-8mm}
\end{figure}

Fig.~\ref{fig:bipartite} shows an example of generated graphs in the database and the ground truth graphs in Task 1. 
The number of nodes on each side is randomly chosen from $[5, 7]$.
They are all complete bipartite graphs.
The parameters are randomly initialized for $50$ times.\looseness=-1

\begin{figure}[h]
\centering
\subfloat[Chain]{\includegraphics[scale=0.4]{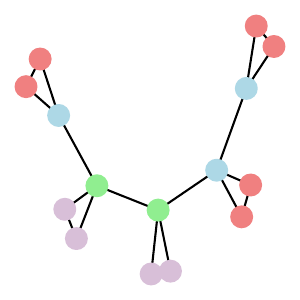}\label{fig:chain1}}
\hspace{2mm}
\subfloat[P1]{\includegraphics[scale=0.4]{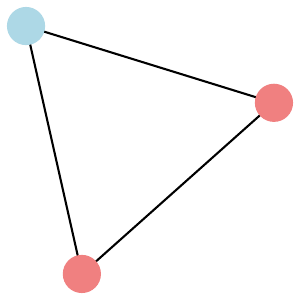}\label{fig:chain2}} 
\hspace{2mm}
\subfloat[P2]{\includegraphics[scale=0.4]{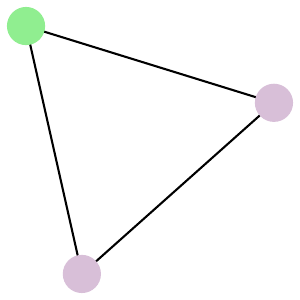}\label{fig:chain3}}
\vspace{-2mm}
\caption{Simple Subgraph Matching in Triangle Chain} \label{fig:chain}
\vspace{-2mm}
\end{figure}

Fig.~\ref{fig:chain} shows an example of generated graphs in the database and the ground truth graphs in Task 2. 
The number of triangles for each chain is randomly chosen from $[3, 5]$.
The color set of each triangle is either red-red-blue (P1, Fig.~\ref{fig:chain2}), or purple-purple-green (P2, Fig.~\ref{fig:chain3}).
The parameters are randomly initialized for $50$ times.\looseness=-1

\begin{figure}[h]
\centering
\captionsetup[subfigure]{width=1in}
\subfloat[3-Regular Graph w/o Colors]{\includegraphics[scale=0.35]{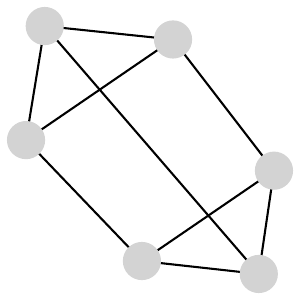}\label{fig:h1}}
\hspace{12mm}
\subfloat[2-Regular Graph w/ Colors]{\includegraphics[scale=0.35]{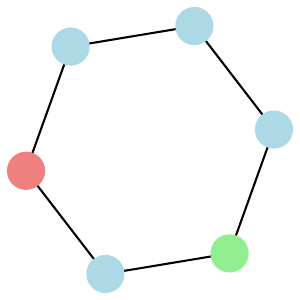}\label{fig:h2}}
\vspace{-2mm}
\caption{Ground Truth Graphs in GED-Based Evaluation} \label{fig:hard}
\vspace{-2mm}
\end{figure}

\subsection{Task 2: GED-Based Evaluation}
We use GED with edge weights, and normalize the edge weights of learned \hiddens into $[0,1]$ before computing GED.
Given an edge with weight $w$, the cost of removing it is $w$, and the cost of fulfilling it is $1 - w$.
The cost of changing the node label is $1$.
In our experiments, we do not need to add or delete node(s).

A sparsity (L1) loss of the \hiddens is used and tuned to prevent learning trivial solutions.
In both Task 2-1 and Task 2-2, the parameters are randomly initialized for $50$ times.
For each of the tasks, we use the same initialization set to run the experiment for several times.
For Task 2-2, the sparsity loss is also adopted on both hidden features, to avoid the local minimums.
Fig.~\ref{fig:hard} shows the ground truth graphs of Task 2 in GED-based evaluation.

\section{Reproducibility} \label{app:rep}

\subsection{Configurations}
The experiments are conducted on a stock Linux server with an NVIDIA RTX A6000 GPU.

\subsection{Search Space of Hyperparameters} \label{app:hyper}

{\footnotesize
\setlength{\tabcolsep}{3pt}
\begin{table}[h]
\caption{Search space of hyperparameters. \label{tab:hyper}}
\vspace{-3mm}
\centering{%\resizebox{0.9\linewidth}{!}{
\begin{tabular}{c | L{5.5cm}}
\hline
\textbf{Application} & \textbf{Hyperparameter Configurations} \\
\hline
Anomaly Detection & Epoch $=200$, lr $=0.001$, \# of Subgraphs $= [8, 16]$, Subgraph Size $= [5, 10]$, \# of Steps $= [2, 3]$ \\
\hline
Substructure Counting & Epoch $=500$, lr $=0.01$, \# of Subgraphs $= 8$, Subgraph Size $= 6$, Egonet Size $= 6$, hop $= 1$, \# of Steps $= [2, 3]$, \# of Layers $= 1$ \\
\hline
Graph Classification & Epoch $=200$, lr $=0.01$, \# of Subgraphs $= 8$, Subgraph Size $= 6$, Egonet Size $= 10$, hop $= 1$, \# of Steps $= [2, 3]$, \# of Layers $= 1$ \\
\hline
\end{tabular}
}%}
\end{table}
}

In Table~\ref{tab:hyper}, we report the search space we use for hyperparameter search in each application in Sec.~\ref{ssec:rwppexp}.
Most hyperparameters follow the default settings in iGAD \cite{zhang2022dual} and KerGNN \cite{feng2022kergnns}, since they are proposed to improve the overall performance.
For substructure counting, in order to show that better extracted features can be used more easily to solve the task, we limit the model to be simple, as it is commonly done in linear probing \cite{alain2017understanding, akhondzadeh2023probing}.
Since the largest degree in the substructure counting dataset is $6$, we set the learnable subgraph size to $6$ as well.
For graph classification, we use the default subgraph and egonet sizes from the original paper.

In Sec.~\ref{ssec:gnnconnection}, we run all the experiments with three random seeds and report the average.
In node classification, for homophily graphs, the random walk step length is set to $4$; for heterophily graphs, TwiBot-22 and graph-level tasks, it is set to $2$.
For node classification and the ogbg-molhiv dataset, the number of layers is set to $2$; for the ZINC and ogbg-pcba datasets, it is set to $6$.

\subsection{Dataset Statistics} \label{app:data}
We report the dataset statistics of each application in Table~\ref{tab:statad}, \ref{tab:statsgc}, \ref{tab:statcls}, \ref{tab:statnc} and \ref{tab:statogb}.
The ``REDDIT'' dataset, which originally includes more than $200$K graphs, is down-sampled for fast evaluation with preserved class ratio.

{\footnotesize
\setlength{\tabcolsep}{3pt}
\begin{table*}[h]
\caption{Dataset statistics of graph anomaly detection. \label{tab:statad}}
\centering{%\resizebox{1\linewidth}{!}{
\begin{tabular}{l | c c c c c c c c c c c}
\hline
\textbf{Dataset} & \textbf{MCF-7} & \textbf{MOLT-4} & \textbf{PC-3} & \textbf{SW-620} & \textbf{NCI-H23} & \textbf{OVCAR-8} & \textbf{P388} & \textbf{SF-295} & \textbf{SN12C} & \textbf{UACC-257} \\
\hline
\# of Graphs & $27,770$ & $39,765$ & $27,509$ & $40,532$ & $40,353$ & $40,516$ & $41,472$ & $40,271$ & $40,004$ & $39,988$ \\
\# of Anomalies & $2,294$ & $3,140$ & $1,568$ & $2,410$ & $2,057$ & $2,079$ & $2,298$ & $2,025$ & $1,955$ & $1,643$ \\
Avg. \# of Nodes & $26.4$ & $26.1$ & $26.4$ & $26.1$ & $26.1$ & $26.1$ & $22.1$ & $26.1$ & $26.1$ & $26.1$ \\
Avg. \# of Edges & $28.5$ & $28.1$ & $28.5$ & $28.1$ & $28.1$ & $28.1$ & $23.6$ & $28.1$ & $28.1$ & $28.1$ \\
\hline
\end{tabular}
}%}
\end{table*}

\begin{table*}[h]
\caption{Dataset statistics of substructure counting. \label{tab:statsgc}}
\centering{%\resizebox{1\linewidth}{!}{
\begin{tabular}{c | C{4cm} c c c c}
\hline
\textbf{Dataset} & \textbf{Task Semantic} & \textbf{\# of Tasks} & \textbf{\# of Graphs} & \textbf{Avg. \# of Nodes} & \textbf{Avg. \# of Edges} \\
\hline
CountingSub. & Normalized number of substructures & $4$ & $1,500/ 1,000 / 2,500$ & $18.8$ & $62.6$ \\
\hline
\end{tabular}
}%} 
\end{table*}

\begin{table*}[h]
\caption{Dataset statistics of graph classification. \label{tab:statcls}}
\centering{%\resizebox{1\linewidth}{!}{
\begin{tabular}{l | c c c c c c c c c c}
\hline
\textbf{Dataset} & \textbf{MUTAG} & \textbf{D\&D} & \textbf{NCI1} & \textbf{PROTEINS} & \textbf{MUTAGEN} & \textbf{TOX21} & \textbf{ENZYMES} & \textbf{IMBD-B} & \textbf{IMDB-M} & \textbf{REDDIT} \\
\hline
\# of Graphs & $188$ & $1,178$ & $4,110$ & $1,113$ & $4,337$ & $8,169$ & $600$ & $1,000$ & $1,500$ & $10,155$ \\
\# of Classes & $2$ & $2$ & $2$ & $2$ & $2$ & $2$ & $6$ & $2$ & $3$ & $2$ \\
Avg. \# of Nodes & $17.9$ & $284.3$ & $29.9$ & $39.1$ & $30.3$ & $18.1$ & $32.6$ & $19.8$ & $13.0$ & $23.9$ \\
Avg. \# of Edges & $19.8$ & $715.7$ & $32.3$ & $72.8$ & $30.8$ & $18.5$ & $62.1$ & $96.5$ & $65.9$ & $25.0$ \\
\hline
\end{tabular}
}%}
\end{table*}
}

{\footnotesize
\setlength{\tabcolsep}{3pt}
\begin{table*}[h]
\caption{Dataset statistics of node classification. \label{tab:statnc}}
\setlength\fboxsep{0pt}
\centering{%\resizebox{1\columnwidth}{!}{
\begin{tabular}{l | cccccc}
\hline
\textbf{Dataset} & \textbf{Cora} & \textbf{CiteSeer} & \textbf{PubMed} & \textbf{Chameleon} & \textbf{Squirrel} & \textbf{Actor} \\
\hline
\# of Nodes & $2,709$ & $3,327$ & $19,717$ & $2,277$ & $5,201$ & $29,926$ \\
\# of Edges & $5,429$ & $4,732$ & $44,338$ & $36,101$ & $216,933$ & $7,600$ \\
\# of Classes & $7$ & $6$ & $3$ & $5$ & $5$ & $5$ \\
\hline
\end{tabular}
}%}
\end{table*}
}

{\footnotesize
\begin{table*}[h]
\caption{Dataset statistics of graph classification and regression. \label{tab:statogb}}
\centering{%\resizebox{1\linewidth}{!}{
\begin{tabular}{l | c c c}
\hline
\textbf{Dataset} & \textbf{ZINC} & \textbf{ogbg-molhiv} & \textbf{ogbg-molpcba} \\
\hline
Task & Regression & Classification & Classification \\
\# of Graphs & $12,000$ & $41,127$ & $437,929$ \\
Avg. \# of Nodes & $17.9$ & $25.5$ & $26.0$ \\
Avg. \# of Edges & $19.8$ & $27.5$ & $28.1$ \\
\hline
\end{tabular}
}%}
\end{table*}
}